\documentclass[letterpaper]{article} 
\usepackage{aaai2026}  
\usepackage{times}  
\usepackage{helvet}  
\usepackage{courier}  
\usepackage[hyphens]{url}  
\usepackage{graphicx} 
\usepackage{natbib}  
\usepackage{caption} 
\usepackage{algorithm}
\usepackage{algorithmic}
\usepackage{multirow}
\usepackage{array} 
\usepackage{booktabs}
\usepackage{makecell}
\usepackage{graphics}

\usepackage{newfloat}
\usepackage{listings}
\DeclareCaptionStyle{ruled}{labelfont=normalfont,labelsep=colon,strut=off} 
\floatstyle{ruled}
\newfloat{listing}{tb}{lst}{}
\floatname{listing}{Listing}
\title{IConv: Focusing on Local Variation with Channel Independent Convolution for Multivariate Time Series Forecasting}
\author {
    Gawon Lee \textsuperscript{\rm 1},
    Hanbyeol Park \textsuperscript{\rm 1},
    Minseop Kim \textsuperscript{\rm 1},
    Dohee Kim \textsuperscript{\rm 1},
    Hyerim Bae \textsuperscript{\rm 1}\thanks{Corresponding Author.}
}
\affiliations {
    \textsuperscript{\rm 1}Pusan National University, Busan, Republic of Korea\\
    \{monago, pb104, rlaals7349, kimdohee, hrbae\}@pusan.ac.kr
}

\begin{document}

\maketitle

\footnote{Submitted to AAAI 2026.}

\begin{abstract}
Real-world time-series data often exhibit non-stationarity, including changing trends, irregular seasonality, and residuals. In terms of changing trends, recently proposed multi-layer perceptron (MLP)-based models have shown excellent performance owing to their computational efficiency and ability to capture long-term dependency. However, the linear nature of MLP architectures poses limitations when applied to channels with diverse distributions, resulting in local variations such as seasonal patterns and residual components being ignored. However, convolutional neural networks (CNNs) can effectively incorporate these variations. To resolve the limitations of MLP, we propose combining them with CNNs. The overall trend is modeled using an MLP to consider long-term dependencies. The CNN uses diverse kernels to model fine-grained local patterns in conjunction with MLP trend predictions. To focus on modeling local variation, we propose IConv, a novel convolutional architecture that processes the temporal dependency channel independently and considers the inter-channel relationship through distinct layers. 
Independent channel processing enables the modeling of diverse local temporal dependencies and the adoption of a large kernel size. Distinct inter-channel considerations reduce computational cost. The proposed model is evaluated through extensive experiments on time-series datasets. The results reveal the superiority of the proposed method for multivariate time-series forecasting. The code is available at https://github.com/qkfxhqrkrrl/IConv.
\end{abstract}

\section{Introduction}

While deep learning has demonstrated remarkable success in multivariate time-series forecasting (MTSF) across applications such as weather forecasting \cite{Weather}, finance \cite{Finance}, and sensors \cite{Sensor}, real-world time-series data often exhibit non-stationarity, presenting complex challenges such as dynamic trends, irregular seasonality, and persistent residuals. To effectively address these complexities, long-term dependencies need to be modeled to capture trends, while concurrently capturing local variation such as seasonal and local temporal fluctuation. The capabilities of multi-layer perceptron (MLP) and transformer-based models are well-established for modeling long-term dependencies. Specifically, MLP-based models have gained traction in recent years owing to their computational efficiency and state-of-the-art prediction performance \cite{HDMixer}. However, the linear nature of MLP poses a limitation when applied to channels with diverse distributions \cite{UMixer}. These models easily overfit to the trend while disregarding local variation, such as shifting seasonality and residuals, as shown in Figure 1. The decomposition of local variation in the image suggests the model is overfitted to a specific channel's local variations. Consequently, MLP-based models exhibit a reduced capacity to capture local fluctuation across the number of channels, thereby decreasing their robustness for nonstationary data. This highlights the need for mechanisms that can capture diverse temporal patterns, which is the core advantage of convolutional neural networks (CNNs).

\begin{figure}[t]
\centering
\includegraphics[width=0.95\columnwidth]{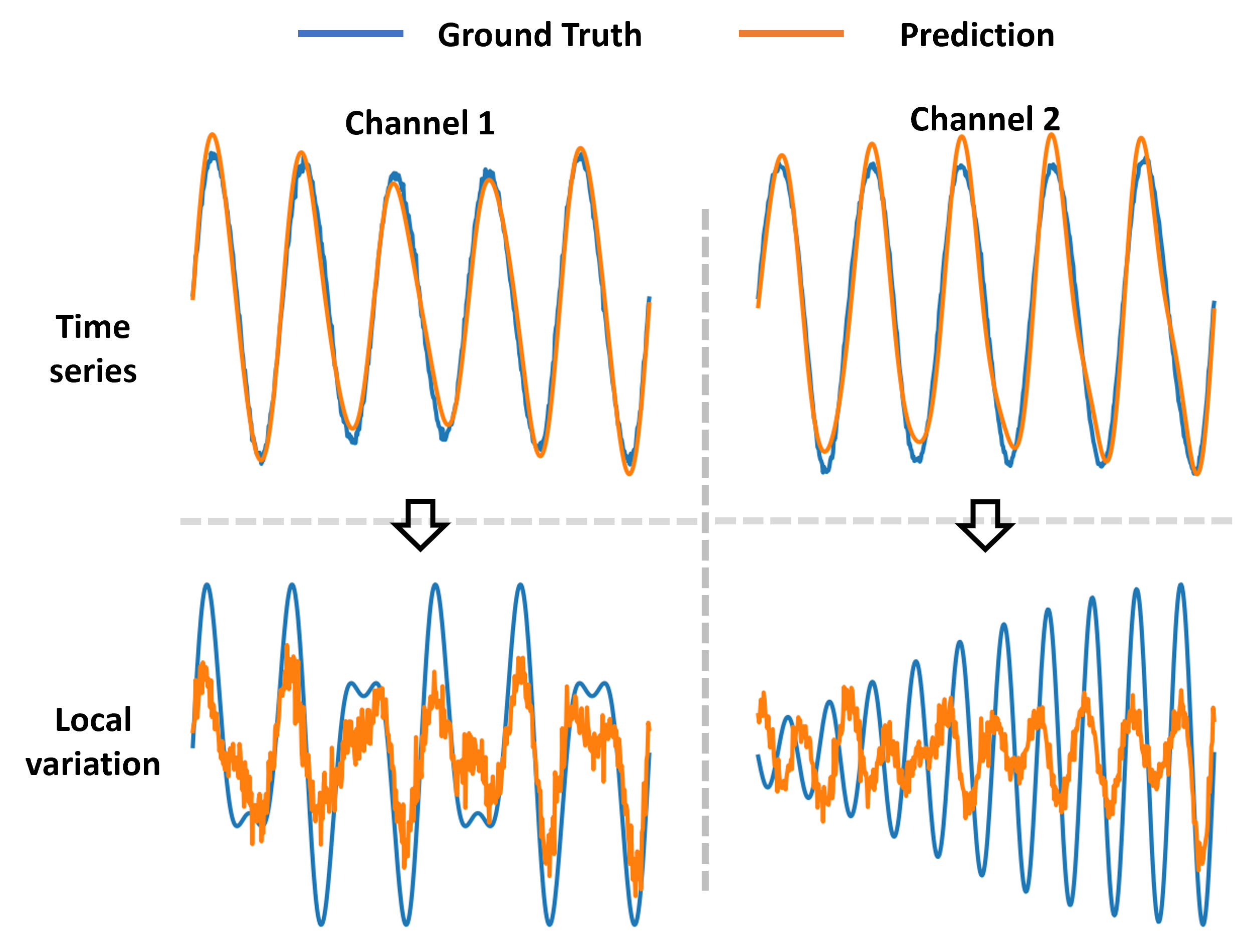} 
\caption{MLP’s failure mode on a synthetic dataset with identical global trends but different local variations.}
\label{fig1}
\end{figure}

CNNs have demonstrated considerable efficacy in capturing temporal patterns within time-series forecasting, particularly showcasing robustness to strong seasonality and other short-term fluctuations \cite{CNN_seasonality,ConvSeasonality}. However, a key limitation of conventional CNNs lies in their constrained receptive field, which restricts effectively modeling long-range dependencies because they struggle to integrate relevant information beyond their defined kernel sizes \cite{CNN_long_term_limit}. While increasing the kernel size might conceptually address this limitation, it leads directly to a substantial increase in computational burden, scaling proportionally with both kernel size and the number of channels. Despite recent advances in CNN-based models that investigate alternative solutions for enhanced efficiency, they generally require more computation than their MLP-based counterparts.
Building on these insights, we propose a novel framework that synergistically combines an MLP with a CNN. In this framework, the MLP predicts the overall macroscopic trends from the input time series. The CNN then models local variations based on the trend predictions made by the MLP. This synergistic combination allows the MLP to robustly capture global dependencies, whereas the CNN effectively models the diverse local temporal patterns that MLPs often miss. Furthermore, to enhance both the efficiency and capacity of the CNN component for capturing the fine-grained local dynamics, we introduced a novel CNN architecture (IConv). IConv decouples the mechanisms of conventional CNN and comprises two key components: a channel independent patcher (CIP) and an inter-channel mixer (ICM). The CIP uses a channel-independent convolution to extract rich local temporal features from each channel. This design considerably reduces the computational requirements, even when using large kernel sizes, and enables the capture of diverse channel-specific representations. The ICM then applies efficient convolutional operations to the extracted channel-independent information, capturing crucial inter-channel relationships without incurring the substantial computational overhead typically associated with standard multi-channel convolutions. The main contributions of the proposed method are summarized as:

\begin{itemize}
\item The combination of MLP and CNN synergistically complements the limitations of each, and contributes to enhancing robustness to non-stationary data.

\item IConv, a specialized convolutional architecture that decomposes standard operations into distinct temporal and inter-channel processes, is introduced. This separation improves computational efficiency substantially and enables larger kernels to be used for richer feature extraction.

\item The proposed IConv-based framework achieves state-of-the-art performance across several MTSF benchmarks.
\end{itemize}

\section{Related Work}

\subsection{MTSF Methodologies}
Early deep learning approaches such as recurrent neural networks struggled with modeling long-term dependencies due to gradient vanishing problems \cite{Gradient_vanishing}. Transformers and recent MLP-based models have addressed these long-term dependency challenges by correlating entire input sequences \cite{iTransformer,Autoformer, DLinear}. However, their global correlation mechanisms often overlook fine-grained local details and struggle to adapt to diverse distributional characteristics across channels, making them vulnerable to non-stationarity \cite{UMixer}. Although methods such as instance normalization \cite{RevIN} and stationary correction \cite{UMixer} partially address non-stationarity, they focus primarily on global trend shifts rather than nuanced local variations. To address this limitation, we utilize CNNs for their proficiency in capturing local temporal patterns and seasonal components by translation-invariant filters \cite{ConvSeasonality}. The research further highlights the capacity of MLP and CNN models to complement each other, generating synergistic effects.

\subsection{CNN in MTSF}
Conventional one-dimensional convolution considers correlation between channels and temporal dependency concurrently \cite{ModernTCN,TimesNet}. However, such an approach increases the computational burden significantly when the number channel increases. Consequently, conventional CNNs struggle to model high-dimensional MTSF data effectively without incurring an excessive computational burden. Despite recent CNN architectures using feedforward networks \cite{ModernTCN} or multi-scaling \cite{MICN} to decrease the computational load, their resource requirements still exceed those of transformer- and MLP-based architectures. Conversely, the proposed model considers the long-term dependency via the MLP, thus eliminating the need to allocate resources to enlarge the receptive field. Given the MLP's focus on global trends, it becomes essential for the CNN component to capture diverse local patterns of individual channels to enhance robustness against the local fluctuations prevalent in non-stationary data \cite{Convolution_inductive_bias}. Recent research indicates that channel-independent MTSF models have demonstrated superior robustness compared to their channel-dependent counterparts, yielding improved predictive accuracy \cite{Channel_independence}. Based on these findings, IConv utilized channel independent convolution to extract more diverse local variation and enhance the robustness of prediction. Furthermore, to compensate for the limitations of channel-independent operations that do not take into account the correlation between channels, we employ $1\times 1$ convolution, the efficacy of which has been established in CV domains \cite{MobileNets}.

\section{Methodology}
The MTSF forecasts a future sequence of length $L$ based on the multivariate input sequence $X_in$ of length $T$. The input can be expressed as $ X_{in}=[X^{(1)}, X^{(2)}, \ldots, X^{(C)}] \in R^{C\times T}$, where $X_{in}^{(c)}=[x_1^{(c)},x_2^{(c)},\ldots, x_T^{(c)}] \in R^{T}$ represents row vectors of the $c$-th channel with length $T$. The model outputs $\hat{Y}=[\hat{Y}^{(1)},\hat{Y}^{(2)},\ldots,\hat{Y}^{(C)}]\in R^{C\times L}$, where $Y^{(c)}=[\hat{x}_{T+1}^{(c)},\hat{x}_{T+2}^{(c)},\ldots,\hat{x}_{T+L}^{(c)}]\in R^L$ is the prediction of the step L future sequence for the c-th channel and $Y\in R^{C\times L}$ is the corresponding ground truth future sequence of every channel. Specifically, for the $c$-th channel, it is $Y^{(c)}=[x_{T+1}^{(c)},x_{T+2}^{(c)},\ldots,x_{T+L}^{(c)}]\in R^L$. The objective of the training is to minimize the gap between the model's predicted output $\hat{Y}$ and the ground truth future sequence $Y$ measured by a loss function. A description of the operations in the IConv block is provided with a visualization of the process in Figure 2. Each component of the figure is explained in this section.

\begin{figure*}[t]
\centering
\includegraphics[width=0.9\textwidth]{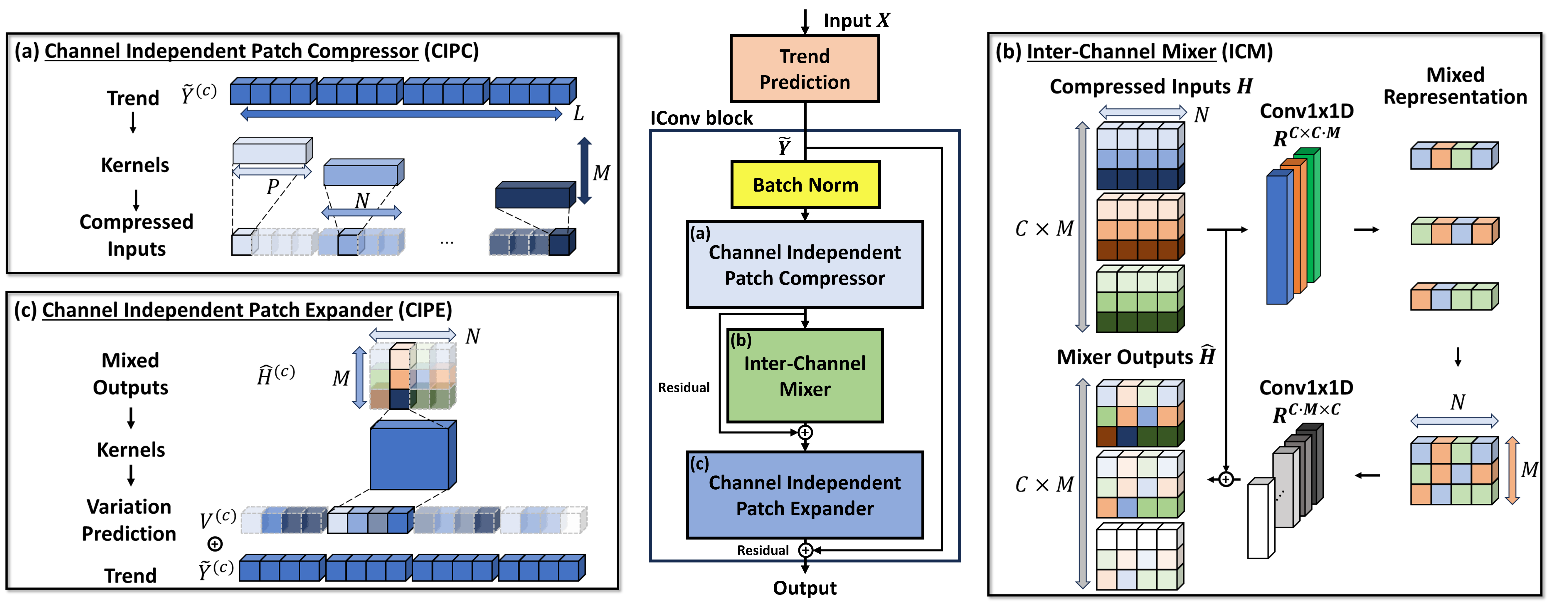} 
\caption{Overall architecture of IConv: (a) CIPC; (b) ICM; (c) CIPE}
\label{fig2}
\end{figure*}

\subsection{Trend Prediction}

Before the input is predicted into a future sequence, $X_{in}$ is encoded using the MLP, which transforms the $T$-length sequence into $d_{model}$-dimensional representation, and then projects it back to a $T$-length representation across every channel. This process is denoted as
\begin{equation}
    \begin{array}{l}
        Enc(X_{in})=ReLU(X_{in}\times W_{p}+b_{p})\times W_{r}+b_{r} \\
        \\
        X_{enc}=X_{in}+Enc(X_{in})
    \end{array}
\end{equation}
where $X_{enc}\in R^{C\times T}$ is the final output of MLP model. The layer that projects to a higher dimension is parameterized by $W_p\in R^{T\times d_model}$ and $b_p\in R^{d_{model}}$, and $d_{model}$ is an adjustable hyperparameter. Parameters $W_r\in R^{d_{model}\times T}$ and $b_r\in R^T$ denote weight and bias of the dimensionality reduction layer, respectively. The output of the encoder is added to the input as a residual, giving $X_{enc}$. Consequently, it is regressed to a future sequence by a single linear layer, and can be denoted as
\begin{equation}
    \tilde{Y}=X_{enc}\times W_{reg}+b_{reg}
\end{equation}
where $\tilde{Y} \in R^{C\times L}$ is the overall trend computed by the trend prediction layer. The function is parameterized by $W_{reg}\in R^{T\times L}$ and $b_{reg}\in R^L$, which project the input sequence into an output sequence.

\subsection{Channel Independent Patch Compressor}

The CIP comprises two main components: CIP compressor (CIPC) and CIP expander (CIPE). The CIPC identifies local temporal dependencies and extracts diverse representations of inputs through channel-independent convolution. The CIPE predicts the local variation through transposed convolution by upsampling the sequential dimension of the hidden states that were compressed by CIPC. The CIPC extracts $M$ representations from each channel through a convolution kernel $K_{cp}^{(c,m)}\in R^P$, where $(c,m)$ is the index of the $m$-th kernel for the $c$-th channel and $P$ is the patch size. The kernel strides each time by $S$; therefore the sequence length after convolution is $N=(L-P)/S+1$, given that the predicted sequence length $L$ and patch size $P$ are completely divisible by stride size $S$ and no padding is applied. CIPC can be expressed as
\begin{equation}
    H^{(c,m)} = ReLU(Conv1D(Norm(\tilde{Y}^{(c)})))_{K_{cp}^{(c,m)}}    
\end{equation}
where $H\in R^{C\cdot M\times N}$ is the output of CIPC and $H^{(c,m)}\in R^N$ is a row vector of outputs computed by convolution kernels $K_{cp}^{(c,m)}$ based on $\tilde{Y}^{(c)}$. The $Norm(\cdot)$ is a batch normalization deployed for the experiments. Each channel is assigned with $K_{cp}^{(c)}=[K_{cp}^{(c,1)},K_{cp}^{(c,2)},\ldots,K_{cp}^{(c,M)}]$ and their computational results are $H^{(c)}=[H^{(c,1)},H^{(c,2)},\ldots,H^{(c,M)}]$. We extract the key features by $ReLU$ instead of pooling to further enhance the efficiency. Owing to the channel-independent design of the CIPC, the convolutional layer's parameter size is reduced to $K\in R^{C\times M\times P}$, where $M$ is an integer typically $< 10$. In contrast, conventional convolutional layers often use kernels of size $K\in R^{C\times Z\times P}$, where the hidden dimension $Z$ can range up to 64. This significant reduction in parameter count contributes to greater efficiency, enabling the use of larger kernel sizes.

\subsection{Inter-Channel Mixer}

To address the constraint of channel independence disregarding the inter-channel correlation, we apply a $1\times 1$ convolution layer, which references the values of other compressed representations across every channel. However, a 1D convolution layer with a kernel size of 1 can be replaced by matrix multiplication along the channel axis. Although the two operations are analogous, using matrix multiplication results in a significant reduction in computation time. To extract the key features from the diverse local temporal representations, ICM first reduces the dimensionality of $H$ to $C$, and then expands it back to $C\cdot M$. The ICM operation can be formalized as:
\begin{equation}
    \bar{H} = ReLU(H\times W_{cr}+b_{cr})
\end{equation}
\begin{equation}
    \hat{H} = ReLU(\bar{H}\times W_{ce}+b_{ce})+\hat{H}
\end{equation}
where $\bar{H}\in R^{C\times N}$ is a compressed mixed representation and $\hat{H}\in R^{C\cdot M\times N}$ is an output of the ICM layer. The parameters for the channel reducing layer are $W_{cr}\in R^{C\times C\cdot M}$ and $b_{cr}\in R^C$, while expansion layers are $W_{ce}\in R^(C\cdot M\times C)$ and $b_{ce}\in R^(C\cdot M)$. The residual connection of $H$ is applied to the output of the expansion layer, preserving the compressed representation and adding extracted information. While each linear layer in the ICM can be replaced by a 1D convolution layer, the results differ slightly owing to different implementation. This is investigated further in the analysis section.

\subsection{Channel Independent Patcher Expander}

The compressed representation following CIPC and ICM comprises embedded hidden states. These states have considered the temporal patterns and interrelationships across varied representations. We upsampled these hidden states along the time axis through a transposed convolution to predict the local variation. This process is:
\begin{equation}
    V_p^{(c)}=TrConv1D(\bar{H}^{(c)})_{K_{ep}^{(c)}} 
\end{equation}
where $K_{ep}^{(c)}\in R^(M\times P)$ represents a kernel of transposed convolution for the $c$-th channel. The input of CIPE $\bar{H}^{(c)}=[\bar{H}{(c,1)},\bar{H}^{(c,2)},\ldots,\bar{H}^{(c,M)}]\in R^{M\times N}$ is a set of $M$ row vectors for the $c$-th channel. These sets of hidden states are considered together using transposed convolution. The output of CIPE is $V\in R^{C\times L}$ and its row vector for the $c$-th channel satisfies $V^{(c)}\in R^L$. This output $V$ is multiplied by variance of $\tilde{Y}$ and added to the prediction of the overall trend $\tilde{Y}$ that was predicted by MLP. This process can be formulated as:  
\begin{equation}
    \hat{Y} = \tilde{Y} + V\times Var(\tilde{Y})	
\end{equation}
where $\hat{Y}\in R^{C\times L}$ is the final prediction made by IConv and $Var(\tilde{Y})\in R^{C\times 1}$ is the variance of each channel along the time axis. The variance is multiplied to local variation prediction V to consider the distributional aspects of each channel, which was lost by the batch normalization in equation (3). As assumed in the CIPC, if the stride and kernel sizes are divisible, the output sequence length matches the original sequence length $L$. The entire process, incorporating CIPC, ICM, and CIPE, can be repeated multiple times by setting the kernel size, stride size, and multiplier for the channel dimensions, as shown in Figure 2. As training progresses, the $V$ values are adjusted to fill the gap that MLPs cannot predict by themselves, and naturally predict the local variation residing in the time-series data.

\subsection{Normalization and Objective Function}

For preprocessing the dataset, we used reversible instance normalization to mitigate the distribution shift between the training and test datasets. The mean and variance of each channel were stored and used to normalize the input sequence. Denormalization was then applied to the prediction sequence using the stored mean and variance. For the training objectives, we set $L_1$ as the loss for its insensitivity to outliers. It is denoted as
\begin{equation}
    Loss=\frac{1}{C}\sum_{c=1}^C|Y^{(c)}-Y^{(c)}|
\end{equation}

\section{Experiments}

\subsection{Datasets}

We evaluated five large-scale real-world time-series datasets for MTSF. (1) \textbf{Electricity consuming load (ECL)} \cite{Autoformer} data comprise hourly electricity consumption with 321 variables; (2) \textbf{Electricity transformer temperature (ETT)} \cite{Autoformer} includes 7 variables resampled using two criteria: hourly sampled data are ETTh1, ETTh2 and ETTm1, and ETTm2 are sampled every 15 min; (3) \textbf{Solar energy} \cite{LTSNet} cite includes 137 variables sampled every 10 mins; (4) \textbf{Traffic} \cite{Autoformer} data describes the occupancy rates of lanes using 862 variables; (5) \textbf{Weather} \cite{Autoformer} includes 21 variables collected every 10 min from a weather station. For all datasets, we divided the data into training, validation, and testing sets in chronological order. The ETT dataset used a 6:2:2 ratio, whereas the remaining datasets were split in a 7:1:2 ratio. 

\subsection{Experimental Setups}
All experiments were conducted in a Google Colab cloud-computing environment using an Nvidia L4 GPU for training. The models were trained for 10 epochs, using the Adam optimizer, with early stopping patience of 3 epochs. To ensure a fair comparison of the convergence rates, a uniform learning rate of 0.001 was initially applied to all models and adjusted via a scheduler after each epoch. Baseline models were selected from state-of-the-art architectures widely used in the MTSF domain. These models can be categorized as: (1) \textbf{Transformer-based models}: iTransformer \cite{iTransformer}, PatchTST \cite{PatchTST}; (2) \textbf{Linear-based models}: TimeMixer++ \cite{TimeMixer++}, Amplifier \cite{Amplifier}, DLinear \cite{DLinear}; and (3) \textbf{CNN-based models}: ModernTCN \cite{ModernTCN}, TimesNet \cite{TimesNet}. After each epoch, the models were evaluated on the validation set using the mean absolute error (MAE). Upon completion of training, the model weights corresponding to the lowest validation loss were loaded and then evaluated on the test dataset using both the mean squared error (MSE) and MAE metrics.


\def\scaleSize{0.8}

\begin{table*}[th]
\begin{small}
\begin{tabular}{cc|cc|cc|cc|cc|cc|cc|cc|cc}

\toprule[1pt]
\multicolumn{2}{c|}{\multirow{2}{*}{Models}} & 
\multicolumn{2}{c|}{\scalebox{0.95}{IConv}}& 
\multicolumn{2}{c|}{\scalebox{0.95}{TimeMixer++}}&
\multicolumn{2}{c|}{\scalebox{0.95}{Amplifier}}&
\multicolumn{2}{c|}{\scalebox{0.95}{ModernTCN}}& 
\multicolumn{2}{c|}{\scalebox{0.95}{iTransformer}}&
\multicolumn{2}{c|}{\scalebox{0.95}{PatchTST}}&
\multicolumn{2}{c|}{\scalebox{0.95}{TimesNet}}&
\multicolumn{2}{c}{\scalebox{0.95}{DLinear}} \\
&
\multicolumn{1}{c|}{\scalebox{0.95}{}}& 
\multicolumn{2}{c|}{\scalebox{0.95}{(Ours)}}& 
\multicolumn{2}{c|}{\scalebox{0.95}{(2025)}}&
\multicolumn{2}{c|}{\scalebox{0.95}{(2025)}}&
\multicolumn{2}{c|}{\scalebox{0.95}{(2024)}}& 
\multicolumn{2}{c|}{\scalebox{0.95}{(2024)}}&
\multicolumn{2}{c|}{\scalebox{0.95}{(2023)}}&
\multicolumn{2}{c|}{\scalebox{0.95}{(2023)}}&
\multicolumn{2}{c}{\scalebox{0.95}{(2023)}}
\\

\bottomrule[0.25pt]

\toprule[0.25pt]
\multicolumn{2}{c|}{\scalebox{0.9}{Metric}}                     
    &\scalebox{\scaleSize}{MSE}&\scalebox{\scaleSize}{MAE}
    &\scalebox{\scaleSize}{MSE}&\scalebox{\scaleSize}{MAE}
    &\scalebox{\scaleSize}{MSE}&\scalebox{\scaleSize}{MAE}
    &\scalebox{\scaleSize}{MSE}&\scalebox{\scaleSize}{MAE}
    &\scalebox{\scaleSize}{MSE}&\scalebox{\scaleSize}{MAE}
    &\scalebox{\scaleSize}{MSE}&\scalebox{\scaleSize}{MAE}
    &\scalebox{\scaleSize}{MSE}&\scalebox{\scaleSize}{MAE}
    &\scalebox{\scaleSize}{MSE}&\scalebox{\scaleSize}{MAE}        
\\
\bottomrule[0.25pt]

\toprule[0.25pt]
\multicolumn{1}{c|}
{\multirow{4}{*}{\rotatebox[origin=c]{90}{ECL}}} 
    &\scalebox{\scaleSize}{96}  
        &\scalebox{\scaleSize}{\textbf{0.140}}&\scalebox{\scaleSize}{\textbf{0.229}}
        &\scalebox{\scaleSize}{0.164}&\scalebox{\scaleSize}{0.252}
        &\scalebox{\scaleSize}{0.154}&\scalebox{\scaleSize}{0.243} 
        &\scalebox{\scaleSize}{\underline{0.153}}&\scalebox{\scaleSize}{0.250}
        &\scalebox{\scaleSize}{0.154}&\scalebox{\scaleSize}{\underline{0.240}}
        &\scalebox{\scaleSize}{0.185}&\scalebox{\scaleSize}{0.258}
        &\scalebox{\scaleSize}{0.167}&\scalebox{\scaleSize}{0.265}
        &\scalebox{\scaleSize}{0.199}&\scalebox{\scaleSize}{0.270}
                
    \\

    \multicolumn{1}{c|}{}
    &\scalebox{\scaleSize}{192}
        &\scalebox{\scaleSize}{\textbf{0.156}}&\scalebox{\scaleSize}{\textbf{0.243}}
        &\scalebox{\scaleSize}{0.180}&\scalebox{\scaleSize}{0.266}
        &\scalebox{\scaleSize}{0.172}&\scalebox{\scaleSize}{0.259}
        &\scalebox{\scaleSize}{\underline{0.167}}&\scalebox{\scaleSize}{0.263}
        &\scalebox{\scaleSize}{\underline{0.167}}&\scalebox{\scaleSize}{\underline{0.252}}
        &\scalebox{\scaleSize}{0.188}&\scalebox{\scaleSize}{0.264}
        &\scalebox{\scaleSize}{0.184}&\scalebox{\scaleSize}{0.279}
        &\scalebox{\scaleSize}{0.197}&\scalebox{\scaleSize}{0.273}

    \\
    \multicolumn{1}{c|}{}                     
    &\scalebox{\scaleSize}{336}
        &\scalebox{\scaleSize}{\textbf{0.167}}&\scalebox{\scaleSize}{\textbf{0.257}}
        &\scalebox{\scaleSize}{0.193}&\scalebox{\scaleSize}{0.277}
        &\scalebox{\scaleSize}{\underline{0.176}}&\scalebox{\scaleSize}{\underline{0.263}}
        &\scalebox{\scaleSize}{\underline{0.176}}&\scalebox{\scaleSize}{0.273}
        &\scalebox{\scaleSize}{0.182}&\scalebox{\scaleSize}{0.268}
        &\scalebox{\scaleSize}{0.203}&\scalebox{\scaleSize}{0.279}
        &\scalebox{\scaleSize}{0.213}&\scalebox{\scaleSize}{0.306}
        &\scalebox{\scaleSize}{0.209}&\scalebox{\scaleSize}{0.288}

    \\
    \multicolumn{1}{c|}{}                     
    &\scalebox{\scaleSize}{720}
        &\scalebox{\scaleSize}{\textbf{0.194}}&\scalebox{\scaleSize}{\textbf{0.281}}
        &\scalebox{\scaleSize}{0.224}&\scalebox{\scaleSize}{0.300}
        &\scalebox{\scaleSize}{0.210}&\scalebox{\scaleSize}{0.293}
        &\scalebox{\scaleSize}{\underline{0.200}}&\scalebox{\scaleSize}{\underline{0.292}}
        &\scalebox{\scaleSize}{0.211}&\scalebox{\scaleSize}{0.293}
        &\scalebox{\scaleSize}{0.244}&\scalebox{\scaleSize}{0.312}
        &\scalebox{\scaleSize}{0.241}&\scalebox{\scaleSize}{0.322}
        &\scalebox{\scaleSize}{0.245}&\scalebox{\scaleSize}{0.320}
    
\\
\bottomrule[0.25pt]

\toprule[0.25pt]
\multicolumn{1}{c|}
{\multirow{4}{*}{\rotatebox[origin=c]{90}{ETTh1}}} 
    &\scalebox{0.9}{96}  
        &\scalebox{\scaleSize}{\textbf{0.377}}&\scalebox{\scaleSize}{\textbf{0.384}}
        &\scalebox{\scaleSize}{0.397}&\scalebox{\scaleSize}{0.398}
        &\scalebox{\scaleSize}{0.385}&\scalebox{\scaleSize}{\underline{0.387}} 
        &\scalebox{\scaleSize}{0.395}&\scalebox{\scaleSize}{0.395}
        &\scalebox{\scaleSize}{0.390}&\scalebox{\scaleSize}{0.399}
        &\scalebox{\scaleSize}{0.384}&\scalebox{\scaleSize}{0.390}
        &\scalebox{\scaleSize}{0.399}&\scalebox{\scaleSize}{0.407}
        &\scalebox{\scaleSize}{\underline{0.380}}&\scalebox{\scaleSize}{\underline{0.387}}
                
    \\

    \multicolumn{1}{c|}{}
    &\scalebox{0.9}{192}
        &\scalebox{\scaleSize}{\underline{0.438}}&\scalebox{\scaleSize}{\textbf{0.417}}
        &\scalebox{\scaleSize}{0.451}&\scalebox{\scaleSize}{0.429}
        &\scalebox{\scaleSize}{0.442}&\scalebox{\scaleSize}{0.421}
        &\scalebox{\scaleSize}{0.462}&\scalebox{\scaleSize}{0.436}
        &\scalebox{\scaleSize}{\underline{0.438}}&\scalebox{\scaleSize}{0.429}
        &\scalebox{\scaleSize}{0.443}&\scalebox{\scaleSize}{0.422}
        &\scalebox{\scaleSize}{0.455}&\scalebox{\scaleSize}{0.440}
        &\scalebox{\scaleSize}{\textbf{0.430}}&\scalebox{\scaleSize}{\underline{0.418}}

    \\
    \multicolumn{1}{c|}{}                     
    &\scalebox{0.9}{336} 
        &\scalebox{\scaleSize}{0.485}&\scalebox{\scaleSize}{\textbf{0.440}}
        &\scalebox{\scaleSize}{0.494}&\scalebox{\scaleSize}{0.447}
        &\scalebox{\scaleSize}{\textbf{0.397}}&\scalebox{\scaleSize}{0.448}
        &\scalebox{\scaleSize}{0.521}&\scalebox{\scaleSize}{0.464}
        &\scalebox{\scaleSize}{0.485}&\scalebox{\scaleSize}{0.452}
        &\scalebox{\scaleSize}{0.485}&\scalebox{\scaleSize}{\underline{0.443}}
        &\scalebox{\scaleSize}{0.516}&\scalebox{\scaleSize}{0.479}
        &\scalebox{\scaleSize}{\underline{0.473}}&\scalebox{\scaleSize}{0.444}

    \\
    \multicolumn{1}{c|}{}                     
    &\scalebox{0.9}{720} 
        &\scalebox{\scaleSize}{\textbf{0.489}}&\scalebox{\scaleSize}{\textbf{0.460}}
        &\scalebox{\scaleSize}{0.498}&\scalebox{\scaleSize}{0.471}
        &\scalebox{\scaleSize}{0.529}&\scalebox{\scaleSize}{0.480}
        &\scalebox{\scaleSize}{0.553}&\scalebox{\scaleSize}{0.508}
        &\scalebox{\scaleSize}{0.498}&\scalebox{\scaleSize}{0.478}
        &\scalebox{\scaleSize}{0.492}&\scalebox{\scaleSize}{\underline{0.465}}
        &\scalebox{\scaleSize}{0.534}&\scalebox{\scaleSize}{0.495}
        &\scalebox{\scaleSize}{\underline{0.491}}&\scalebox{\scaleSize}{0.488}
				
\\
\bottomrule[0.25pt]

\toprule[0.25pt]
\multicolumn{1}{c|}
{\multirow{4}{*}{\rotatebox[origin=c]{90}{ETTh2}}} 
    &\scalebox{0.9}{96}  
        &\scalebox{\scaleSize}{0.290}&\scalebox{\scaleSize}{\underline{0.332}}
        &\scalebox{\scaleSize}{0.295}&\scalebox{\scaleSize}{0.338}
        &\scalebox{\scaleSize}{0.285}&\scalebox{\scaleSize}{\underline{0.332}} 
        &\scalebox{\scaleSize}{\textbf{0.280}}&\scalebox{\scaleSize}{0.335}
        &\scalebox{\scaleSize}{0.293}&\scalebox{\scaleSize}{0.339}
        &\scalebox{\scaleSize}{\underline{0.281}}&\scalebox{\scaleSize}{\textbf{0.329}}
        &\scalebox{\scaleSize}{0.315}&\scalebox{\scaleSize}{0.355}
        &\scalebox{\scaleSize}{0.293}&\scalebox{\scaleSize}{0.344}
                
    \\

    \multicolumn{1}{c|}{}
    &\scalebox{0.9}{192}
        &\scalebox{\scaleSize}{\textbf{0.359}}&\scalebox{\scaleSize}{\textbf{0.379}}
        &\scalebox{\scaleSize}{0.371}&\scalebox{\scaleSize}{0.386}
        &\scalebox{\scaleSize}{0.362}&\scalebox{\scaleSize}{\underline{0.382}}
        &\scalebox{\scaleSize}{0.373}&\scalebox{\scaleSize}{0.387}
        &\scalebox{\scaleSize}{0.377}&\scalebox{\scaleSize}{0.392}
        &\scalebox{\scaleSize}{\underline{0.360}}&\scalebox{\scaleSize}{\textbf{0.379}}
        &\scalebox{\scaleSize}{0.383}&\scalebox{\scaleSize}{0.397}
        &\scalebox{\scaleSize}{0.377}&\scalebox{\scaleSize}{0.396}

    \\
    \multicolumn{1}{c|}{}                     
    &\scalebox{0.9}{336} 
        &\scalebox{\scaleSize}{\underline{0.407}}&\scalebox{\scaleSize}{\textbf{0.415}}
        &\scalebox{\scaleSize}{0.422}&\scalebox{\scaleSize}{0.426}
        &\scalebox{\scaleSize}{0.415}&\scalebox{\scaleSize}{0.420}
        &\scalebox{\scaleSize}{\underline{0.407}}&\scalebox{\scaleSize}{\underline{0.418}}
        &\scalebox{\scaleSize}{0.438}&\scalebox{\scaleSize}{0.433}
        &\scalebox{\scaleSize}{\textbf{0.404}}&\scalebox{\scaleSize}{\textbf{0.415}}
        &\scalebox{\scaleSize}{0.447}&\scalebox{\scaleSize}{0.434}
        &\scalebox{\scaleSize}{0.449}&\scalebox{\scaleSize}{0.451}

    \\
    \multicolumn{1}{c|}{}                     
    &\scalebox{0.9}{720} 
        &\scalebox{\scaleSize}{\textbf{0.410}}&\scalebox{\scaleSize}{\textbf{0.431}}
        &\scalebox{\scaleSize}{0.430}&\scalebox{\scaleSize}{0.441}
        &\scalebox{\scaleSize}{0.427}&\scalebox{\scaleSize}{0.439}
        &\scalebox{\scaleSize}{0.413}&\scalebox{\scaleSize}{0.435}
        &\scalebox{\scaleSize}{0.438}&\scalebox{\scaleSize}{0.447}
        &\scalebox{\scaleSize}{\underline{0.412}}&\scalebox{\scaleSize}{\textbf{0.431}}
        &\scalebox{\scaleSize}{0.422}&\scalebox{\scaleSize}{0.436}
        &\scalebox{\scaleSize}{0.610}&\scalebox{\scaleSize}{0.548}
	
\\
\bottomrule[0.25pt]

\toprule[0.25pt]
\multicolumn{1}{c|}
{\multirow{4}{*}{\rotatebox[origin=c]{90}{ETTm1}}} 
    &\scalebox{0.9}{96}  
        &\scalebox{\scaleSize}{\textbf{0.304}}&\scalebox{\scaleSize}{\textbf{0.332}}
        &\scalebox{\scaleSize}{0.315}&\scalebox{\scaleSize}{0.343}
        &\scalebox{\scaleSize}{0.316}&\scalebox{\scaleSize}{0.342} 
        &\scalebox{\scaleSize}{0.342}&\scalebox{\scaleSize}{0.377}
        &\scalebox{\scaleSize}{0.315}&\scalebox{\scaleSize}{0.345}
        &\scalebox{\scaleSize}{\underline{0.312}}&\scalebox{\scaleSize}{\underline{0.338}}
        &\scalebox{\scaleSize}{0.394}&\scalebox{\scaleSize}{0.389}
        &\scalebox{\scaleSize}{0.332}&\scalebox{\scaleSize}{0.351}
                
    \\

    \multicolumn{1}{c|}{}
    &\scalebox{0.9}{192}
        &\scalebox{\scaleSize}{\textbf{0.357}}&\scalebox{\scaleSize}{\textbf{0.361}}
        &\scalebox{\scaleSize}{0.363}&\scalebox{\scaleSize}{0.369}
        &\scalebox{\scaleSize}{0.369}&\scalebox{\scaleSize}{0.367}
        &\scalebox{\scaleSize}{0.373}&\scalebox{\scaleSize}{0.396}
        &\scalebox{\scaleSize}{0.366}&\scalebox{\scaleSize}{0.370}
        &\scalebox{\scaleSize}{\underline{0.361}}&\scalebox{\scaleSize}{\underline{0.365}}
        &\scalebox{\scaleSize}{0.452}&\scalebox{\scaleSize}{0.426}
        &\scalebox{\scaleSize}{0.376}&\scalebox{\scaleSize}{0.374}

    \\
    \multicolumn{1}{c|}{}                     
    &\scalebox{0.9}{336} 
        &\scalebox{\scaleSize}{\underline{0.391}}&\scalebox{\scaleSize}{\textbf{0.384}}
        &\scalebox{\scaleSize}{\textbf{0.386}}&\scalebox{\scaleSize}{0.391}
        &\scalebox{\scaleSize}{0.401}&\scalebox{\scaleSize}{0.390}
        &\scalebox{\scaleSize}{0.420}&\scalebox{\scaleSize}{0.422}
        &\scalebox{\scaleSize}{0.405}&\scalebox{\scaleSize}{0.395}
        &\scalebox{\scaleSize}{0.395}&\scalebox{\scaleSize}{\underline{0.389}}
        &\scalebox{\scaleSize}{0.471}&\scalebox{\scaleSize}{0.439}
        &\scalebox{\scaleSize}{0.406}&\scalebox{\scaleSize}{0.395}

    \\
    \multicolumn{1}{c|}{}                     
    &\scalebox{0.9}{720} 
        &\scalebox{\scaleSize}{\textbf{0.457}}&\scalebox{\scaleSize}{\textbf{0.423}}
        &\scalebox{\scaleSize}{0.459}&\scalebox{\scaleSize}{0.432}
        &\scalebox{\scaleSize}{0.472}&\scalebox{\scaleSize}{0.430}
        &\scalebox{\scaleSize}{0.472}&\scalebox{\scaleSize}{0.430}
        &\scalebox{\scaleSize}{0.463}&\scalebox{\scaleSize}{0.445}
        &\scalebox{\scaleSize}{0.471}&\scalebox{\scaleSize}{0.433}
        &\scalebox{\scaleSize}{\underline{0.458}}&\scalebox{\scaleSize}{\underline{0.427}}
        &\scalebox{\scaleSize}{0.538}&\scalebox{\scaleSize}{0.476}
\\
\bottomrule[0.25pt]

\toprule[0.25pt]
\multicolumn{1}{c|}
{\multirow{4}{*}{\rotatebox[origin=c]{90}{ETTm2}}} 
    &\scalebox{0.9}{96}  
        &\scalebox{\scaleSize}{\textbf{0.168}}&\scalebox{\scaleSize}{\textbf{0.246}}
        &\scalebox{\scaleSize}{0.175}&\scalebox{\scaleSize}{0.251}
        &\scalebox{\scaleSize}{0.171}&\scalebox{\scaleSize}{\underline{0.249}} 
        &\scalebox{\scaleSize}{\underline{0.169}}&\scalebox{\scaleSize}{0.250}
        &\scalebox{\scaleSize}{0.174}&\scalebox{\scaleSize}{0.251}
        &\scalebox{\scaleSize}{0.177}&\scalebox{\scaleSize}{0.253}
        &\scalebox{\scaleSize}{0.187}&\scalebox{\scaleSize}{0.261}
        &\scalebox{\scaleSize}{0.183}&\scalebox{\scaleSize}{0.257}
                
    \\

    \multicolumn{1}{c|}{}
    &\scalebox{0.9}{192}
        &\scalebox{\scaleSize}{\textbf{0.228}}&\scalebox{\scaleSize}{\textbf{0.288}}
        &\scalebox{\scaleSize}{0.245}&\scalebox{\scaleSize}{0.298}
        &\scalebox{\scaleSize}{\underline{0.234}}&\scalebox{\scaleSize}{\underline{0.292}}
        &\scalebox{\scaleSize}{0.240}&\scalebox{\scaleSize}{0.299}
        &\scalebox{\scaleSize}{0.243}&\scalebox{\scaleSize}{0.298}
        &\scalebox{\scaleSize}{0.242}&\scalebox{\scaleSize}{0.296}
        &\scalebox{\scaleSize}{0.260}&\scalebox{\scaleSize}{0.307}
        &\scalebox{\scaleSize}{0.245}&\scalebox{\scaleSize}{0.302}

    \\
    \multicolumn{1}{c|}{}                     
    &\scalebox{0.9}{336} 
        &\scalebox{\scaleSize}{\textbf{0.288}}&\scalebox{\scaleSize}{\textbf{0.326}}
        &\scalebox{\scaleSize}{\underline{0.296}}&\scalebox{\scaleSize}{\underline{0.332}}
        &\scalebox{\scaleSize}{0.300}&\scalebox{\scaleSize}{0.334}
        &\scalebox{\scaleSize}{\underline{0.296}}&\scalebox{\scaleSize}{0.337}
        &\scalebox{\scaleSize}{0.305}&\scalebox{\scaleSize}{0.337}
        &\scalebox{\scaleSize}{0.302}&\scalebox{\scaleSize}{0.337}
        &\scalebox{\scaleSize}{0.322}&\scalebox{\scaleSize}{0.346}
        &\scalebox{\scaleSize}{0.307}&\scalebox{\scaleSize}{0.348}

    \\
    \multicolumn{1}{c|}{}                     
    &\scalebox{0.9}{720} 
        &\scalebox{\scaleSize}{\underline{0.397}}&\scalebox{\scaleSize}{\underline{0.390}}
        &\scalebox{\scaleSize}{0.440}&\scalebox{\scaleSize}{0.418}
        &\scalebox{\scaleSize}{\textbf{0.391}}&\scalebox{\scaleSize}{\textbf{0.389}}
        &\scalebox{\scaleSize}{0.453}&\scalebox{\scaleSize}{0.424}
        &\scalebox{\scaleSize}{0.403}&\scalebox{\scaleSize}{0.394}
        &\scalebox{\scaleSize}{0.404}&\scalebox{\scaleSize}{0.395}
        &\scalebox{\scaleSize}{0.431}&\scalebox{\scaleSize}{0.407}
        &\scalebox{\scaleSize}{0.413}&\scalebox{\scaleSize}{0.419}

\\
\bottomrule[0.25pt]

\toprule[0.25pt]
\multicolumn{1}{c|}
{\multirow{4}{*}{\rotatebox[origin=c]{90}{Solar}}} 
    &\scalebox{\scaleSize}{96}  
        &\scalebox{\scaleSize}{0.194}&\scalebox{\scaleSize}{0.209}
        &\scalebox{\scaleSize}{\textbf{0.184}}&\scalebox{\scaleSize}{\textbf{0.207}}
        &\scalebox{\scaleSize}{\underline{0.185}}&\scalebox{\scaleSize}{\underline{0.208}} 
        &\scalebox{\scaleSize}{0.206}&\scalebox{\scaleSize}{0.233}
        &\scalebox{\scaleSize}{0.200}&\scalebox{\scaleSize}{0.209}
        &\scalebox{\scaleSize}{0.219}&\scalebox{\scaleSize}{0.230}
        &\scalebox{\scaleSize}{0.230}&\scalebox{\scaleSize}{0.227}
        &\scalebox{\scaleSize}{0.286}&\scalebox{\scaleSize}{0.294}
                
    \\

    \multicolumn{1}{c|}{}
    &\scalebox{\scaleSize}{192}
        &\scalebox{\scaleSize}{0.237}&\scalebox{\scaleSize}{\textbf{0.232}}
        &\scalebox{\scaleSize}{\textbf{0.231}}&\scalebox{\scaleSize}{0.236}
        &\scalebox{\scaleSize}{\underline{0.234}}&\scalebox{\scaleSize}{0.235}
        &\scalebox{\scaleSize}{0.241}&\scalebox{\scaleSize}{0.252}
        &\scalebox{\scaleSize}{0.236}&\scalebox{\scaleSize}{\underline{0.233}}
        &\scalebox{\scaleSize}{0.249}&\scalebox{\scaleSize}{0.249}
        &\scalebox{\scaleSize}{0.271}&\scalebox{\scaleSize}{0.252}
        &\scalebox{\scaleSize}{0.318}&\scalebox{\scaleSize}{0.313}

    \\
    \multicolumn{1}{c|}{}                     
    &\scalebox{\scaleSize}{336} 
        &\scalebox{\scaleSize}{0.259}&\scalebox{\scaleSize}{\underline{0.245}}
        &\scalebox{\scaleSize}{0.261}&\scalebox{\scaleSize}{0.255}
        &\scalebox{\scaleSize}{\textbf{0.244}}&\scalebox{\scaleSize}{\textbf{0.243}}
        &\scalebox{\scaleSize}{0.264}&\scalebox{\scaleSize}{0.265}
        &\scalebox{\scaleSize}{\underline{0.253}}&\scalebox{\scaleSize}{0.248}
        &\scalebox{\scaleSize}{0.274}&\scalebox{\scaleSize}{0.263}
        &\scalebox{\scaleSize}{0.305}&\scalebox{\scaleSize}{0.281}
        &\scalebox{\scaleSize}{0.363}&\scalebox{\scaleSize}{0.327}

    \\
    \multicolumn{1}{c|}{}                     
    &\scalebox{\scaleSize}{720} 
        &\scalebox{\scaleSize}{\underline{0.262}}&\scalebox{\scaleSize}{\underline{0.248}}
        &\scalebox{\scaleSize}{0.263}&\scalebox{\scaleSize}{0.256}
        &\scalebox{\scaleSize}{\textbf{0.257}}&\scalebox{\scaleSize}{0.249}
        &\scalebox{\scaleSize}{0.267}&\scalebox{\scaleSize}{0.266}
        &\scalebox{\scaleSize}{\textbf{0.257}}&\scalebox{\scaleSize}{\textbf{0.246}}
        &\scalebox{\scaleSize}{0.278}&\scalebox{\scaleSize}{0.263}
        &\scalebox{\scaleSize}{0.308}&\scalebox{\scaleSize}{0.277}
        &\scalebox{\scaleSize}{0.374}&\scalebox{\scaleSize}{0.322}
    
\\
\bottomrule[0.25pt]

\toprule[0.25pt]
\multicolumn{1}{c|}
{\multirow{4}{*}{\rotatebox[origin=c]{90}{Traffic}}} 
    &\scalebox{\scaleSize}{96}  
        &\scalebox{\scaleSize}{0.459}&\scalebox{\scaleSize}{\textbf{0.263}}
        &\scalebox{\scaleSize}{0.461}&\scalebox{\scaleSize}{\textbf{0.263}}
        &\scalebox{\scaleSize}{0.447}&\scalebox{\scaleSize}{0.283} 
        &\scalebox{\scaleSize}{\underline{0.456}}&\scalebox{\scaleSize}{0.289}
        &\scalebox{\scaleSize}{\textbf{0.408}}&\scalebox{\scaleSize}{\underline{0.268}}
        &\scalebox{\scaleSize}{0.499}&\scalebox{\scaleSize}{0.291}
        &\scalebox{\scaleSize}{0.622}&\scalebox{\scaleSize}{0.312}
        &\scalebox{\scaleSize}{0.671}&\scalebox{\scaleSize}{0.366}
                
    \\

    \multicolumn{1}{c|}{}
    &\scalebox{\scaleSize}{192}
        &\scalebox{\scaleSize}{0.466}&\scalebox{\scaleSize}{\textbf{0.271}}
        &\scalebox{\scaleSize}{0.474}&\scalebox{\scaleSize}{\underline{0.272}}
        &\scalebox{\scaleSize}{\underline{0.457}}&\scalebox{\scaleSize}{0.285}
        &\scalebox{\scaleSize}{0.460}&\scalebox{\scaleSize}{0.286}
        &\scalebox{\scaleSize}{\textbf{0.430}}&\scalebox{\scaleSize}{0.276}
        &\scalebox{\scaleSize}{0.496}&\scalebox{\scaleSize}{0.291}
        &\scalebox{\scaleSize}{0.642}&\scalebox{\scaleSize}{0.319}
        &\scalebox{\scaleSize}{0.625}&\scalebox{\scaleSize}{0.343}

    \\
    \multicolumn{1}{c|}{}                     
    &\scalebox{\scaleSize}{336} 
        &\scalebox{\scaleSize}{0.479}&\scalebox{\scaleSize}{\textbf{0.274}}
        &\scalebox{\scaleSize}{0.474}&\scalebox{\scaleSize}{\underline{0.279}}
        &\scalebox{\scaleSize}{\underline{0.471}}&\scalebox{\scaleSize}{0.292}
        &\scalebox{\scaleSize}{0.475}&\scalebox{\scaleSize}{0.292}
        &\scalebox{\scaleSize}{\textbf{0.446}}&\scalebox{\scaleSize}{0.281}
        &\scalebox{\scaleSize}{0.512}&\scalebox{\scaleSize}{0.297}
        &\scalebox{\scaleSize}{0.652}&\scalebox{\scaleSize}{0.326}
        &\scalebox{\scaleSize}{0.629}&\scalebox{\scaleSize}{0.345}

    \\
    \multicolumn{1}{c|}{}                     
    &\scalebox{\scaleSize}{720} 
        &\scalebox{\scaleSize}{0.517}&\scalebox{\scaleSize}{\textbf{0.293}}
        &\scalebox{\scaleSize}{0.547}&\scalebox{\scaleSize}{\underline{0.299}}
        &\scalebox{\scaleSize}{\underline{0.506}}&\scalebox{\scaleSize}{0.314}
        &\scalebox{\scaleSize}{0.511}&\scalebox{\scaleSize}{0.311}
        &\scalebox{\scaleSize}{\textbf{0.481}}&\scalebox{\scaleSize}{0.300}
        &\scalebox{\scaleSize}{0.558}&\scalebox{\scaleSize}{0.323}
        &\scalebox{\scaleSize}{0.698}&\scalebox{\scaleSize}{0.349}
        &\scalebox{\scaleSize}{0.660}&\scalebox{\scaleSize}{0.365}
    	
\\
\bottomrule[0.25pt]

\toprule[0.25pt]
\multicolumn{1}{c|}
{\multirow{4}{*}{\rotatebox[origin=c]{90}{Weather}}} 
    &\scalebox{0.9}{96}  
        &\scalebox{\scaleSize}{\textbf{0.152}}&\scalebox{\scaleSize}{\textbf{0.189}}
        &\scalebox{\scaleSize}{0.162}&\scalebox{\scaleSize}{0.197}
        &\scalebox{\scaleSize}{\underline{0.153}}&\scalebox{\scaleSize}{\underline{0.190}} 
        &\scalebox{\scaleSize}{\textbf{0.152}}&\scalebox{\scaleSize}{0.194}
        &\scalebox{\scaleSize}{0.210}&\scalebox{\scaleSize}{0.240}
        &\scalebox{\scaleSize}{0.173}&\scalebox{\scaleSize}{0.205}
        &\scalebox{\scaleSize}{0.162}&\scalebox{\scaleSize}{0.204}
        &\scalebox{\scaleSize}{0.208}&\scalebox{\scaleSize}{0.232}
                
    \\

    \multicolumn{1}{c|}{}
    &\scalebox{0.9}{192}
        &\scalebox{\scaleSize}{\textbf{0.201}}&\scalebox{\scaleSize}{\textbf{0.236}}
        &\scalebox{\scaleSize}{0.211}&\scalebox{\scaleSize}{0.243}
        &\scalebox{\scaleSize}{0.205}&\scalebox{\scaleSize}{\underline{0.240}}
        &\scalebox{\scaleSize}{\underline{0.204}}&\scalebox{\scaleSize}{0.242}
        &\scalebox{\scaleSize}{0.264}&\scalebox{\scaleSize}{0.286}
        &\scalebox{\scaleSize}{0.219}&\scalebox{\scaleSize}{0.247}
        &\scalebox{\scaleSize}{0.215}&\scalebox{\scaleSize}{0.252}
        &\scalebox{\scaleSize}{0.245}&\scalebox{\scaleSize}{0.267}

    \\
    \multicolumn{1}{c|}{}                     
    &\scalebox{0.9}{336} 
        &\scalebox{\scaleSize}{\textbf{0.258}}&\scalebox{\scaleSize}{\textbf{0.280}}
        &\scalebox{\scaleSize}{0.269}&\scalebox{\scaleSize}{\underline{0.286}}
        &\scalebox{\scaleSize}{0.267}&\scalebox{\scaleSize}{\underline{0.286}}
        &\scalebox{\scaleSize}{\underline{0.263}}&\scalebox{\scaleSize}{0.289}
        &\scalebox{\scaleSize}{0.310}&\scalebox{\scaleSize}{0.320}
        &\scalebox{\scaleSize}{0.275}&\scalebox{\scaleSize}{0.287}
        &\scalebox{\scaleSize}{0.295}&\scalebox{\scaleSize}{0.306}
        &\scalebox{\scaleSize}{0.288}&\scalebox{\scaleSize}{0.307}

    \\
    \multicolumn{1}{c|}{}                     
    &\scalebox{0.9}{720} 
        &\scalebox{\scaleSize}{\textbf{0.338}}&\scalebox{\scaleSize}{\textbf{0.332}}
        &\scalebox{\scaleSize}{0.352}&\scalebox{\scaleSize}{0.341}
        &\scalebox{\scaleSize}{\underline{0.344}}&\scalebox{\scaleSize}{\underline{0.338}}
        &\scalebox{\scaleSize}{\underline{0.344}}&\scalebox{\scaleSize}{0.342}
        &\scalebox{\scaleSize}{0.377}&\scalebox{\scaleSize}{0.362}
        &\scalebox{\scaleSize}{0.353}&\scalebox{\scaleSize}{0.338}
        &\scalebox{\scaleSize}{0.386}&\scalebox{\scaleSize}{0.369}
        &\scalebox{\scaleSize}{0.347}&\scalebox{\scaleSize}{0.358}
    	
\\
\bottomrule[1pt]

\\            
\end{tabular}
\end{small}
\caption{MTSF forecasting results evaluated by MSE and MAE. The input window was set to 96 and the prediction length varied by {96, 192, 336, 720}. The best results are highlighted in bold, and the second best are underlined.}
\end{table*}

\subsection{Model Configuration}

The MLP encoder architecture, specifically the number of layers and hidden dimension size, varies by the dataset size and characteristics of them. For local variation prediction, three IConv layers were used, each with a uniform stride but with distinct kernel sizes. The kernel sizes were progressively reduced as the IConv layers became shallower, starting with larger kernels in the first layer to capture extensive temporal dependencies and then decreasing them to extract more specific patterns. The combination of kernel sizes is determined by the multiplication of the smallest size; for example, if the smallest kernel is 4, the kernel size is set to $P=[12,8,4]$. Stride sizes were chosen based on the time interval of the dataset. For example, in the case of the dataset sampled every 15 minutes, the stride size was set to 4; conversely, a stride of 3 was applied to the dataset sampled every 10 minutes.

\subsection{Result and Analysis}
Table 1 presents a quantitative evaluation of the proposed IConv method in comparison with existing baseline models. Our IConv consistently outperforms these baselines across several datasets and prediction sequence settings, achieving 45 first-place out of 64 evaluations and rankings second-place in 9 cases. This demonstrates the superior performance of IConv in MTSF. The effectiveness of our model was further validated by visual inspection of the prediction results, as shown in Figure 3. The prediction at the top most is the result of the regression layer. The MLP layer outputs merely follow a recurring macroscopic pattern, missing these crucial local variations. The predictions on the left side, marked by red lines, each represent the output of three IConv layer with each kernel size of $P=[24,16,8]$ with stride size $S=4$. The forecasts of each layer reduce the distance between the MLP outputs and the ground truth, as depicted in the right-hand graphs of the figure. Within the highlighted regions, IConv significantly reduced prediction errors by effectively modeling local temporal patterns. These findings strongly support our core proposition that IConv is highly effective for modeling fined-grained local temporal patterns within non-stationary time-series data.

\begin{figure}[t]
\centering
\includegraphics[width=0.95\columnwidth]{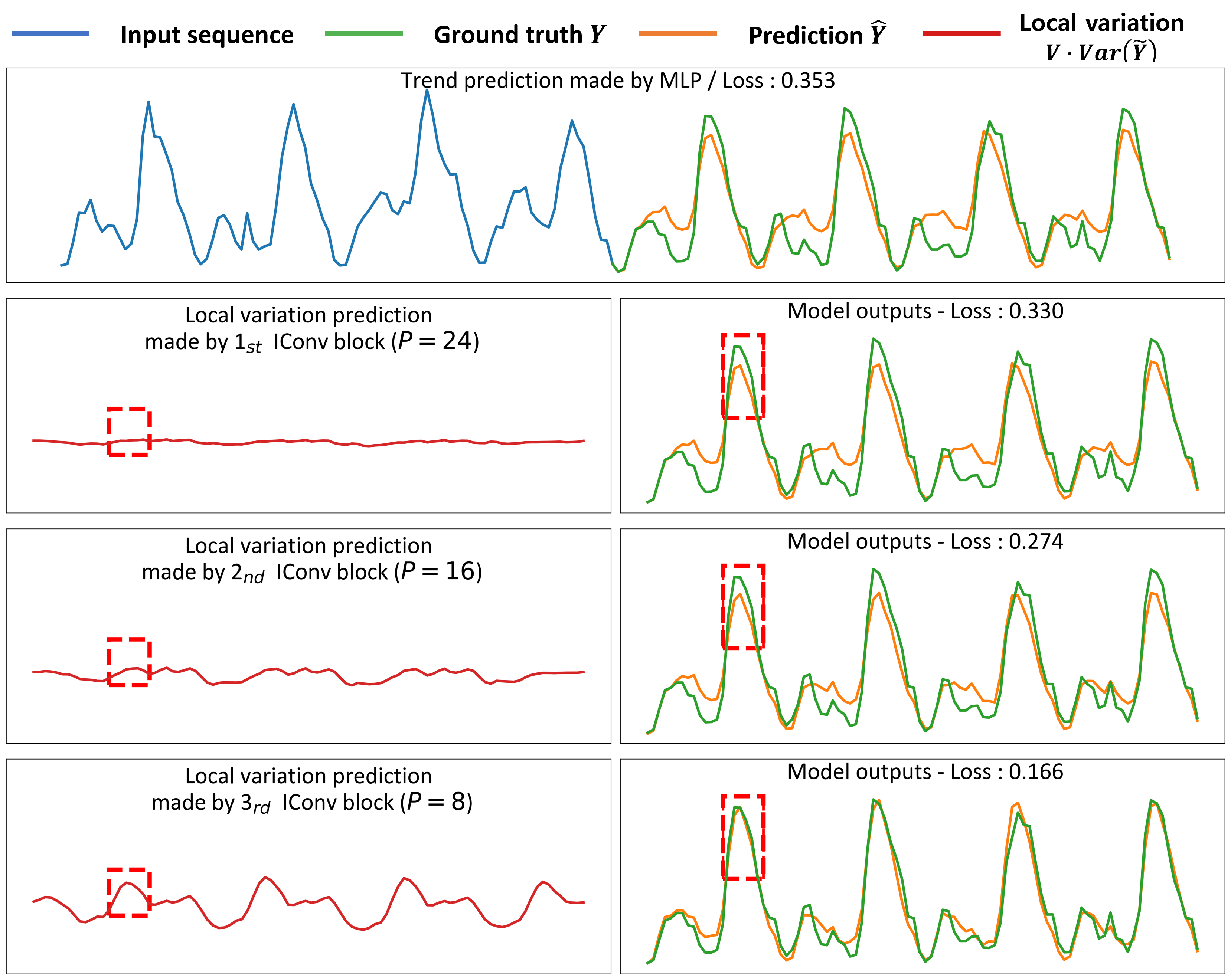} 
\caption{Model results on ECL dataset. Top: Output of regression layer: Input – Blue, future – Green, predictions - Orange; Left: Local variation predictions; Right; Outputs}
\label{fig3}
\end{figure}

\subsection{Receptive Field Analysis}
To further validate the contribution of IConv in capturing local variations, we visualized its receptive field, presented in Figure 4. Receptive fields are generally used in convolutional models to assess the propagation of information through deeper layers, thus they are not directly related to prediction accuracy, and the values of heatmaps might not be a suitable metric for evaluating the input utilization of MLP-based models. However, the comparison with IConvs and MLP help to identify the role of IConv's within our MLP-CNN framework. Following the methodology of \cite{DeadPixel}, we sampled 50 instances of inputs each with 96 timesteps from the ECL dataset. We then computed the gradients of the values at the 48th predicted time step with respect to the input values; the heatmap values reflects each input value's contribution to prediction of middle of the output sequence. The IConv model's capacity to manage localized temporal relationships is indicated by the more dispersed heatmap values relative to the MLP model's heatmap. While MLP reference single timesteps on a 24-hour cycle, IConv leverages contextual information beyond single timestep. Furthermore, there are distinct difference exists in receptive field of ModernTCN and IConv. Within IConv's receptive field, there is significant periodicity and it accounts for long-term dependency by referencing both the initial and final input indices. In contrast, ModernTCN's gradient values are concentrated in the middle area, and the impact diminishes as it extends to the sides.

\begin{figure}[th]
\centering
\includegraphics[width=0.95\columnwidth]{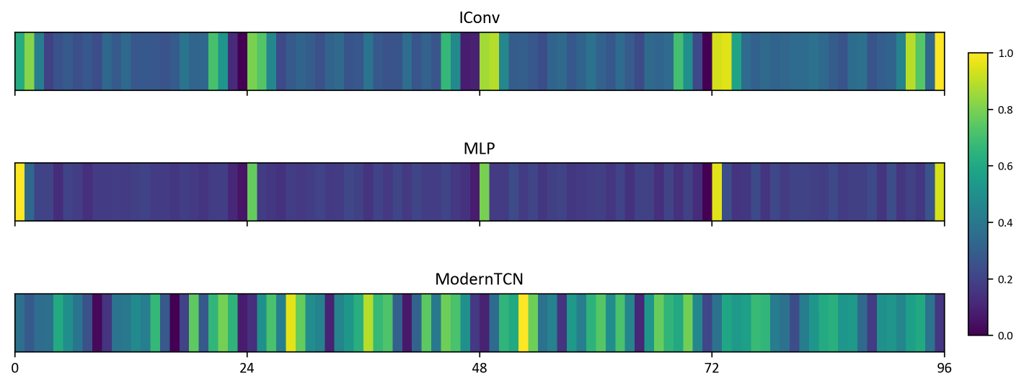} 
\caption{Visualization of receptive field from IConv, MLP and ModernTCN on ECL dataset}
\label{fig4}
\end{figure}

\subsection{Ablation Study}

An ablation study was conducted to quantify the individual contributions of each component within the proposed model. We defined the following variants: \textbf{W/O IConv} is a simple MLP model with RevIN. \textbf{W/O ICM} is an MLP in addition with CIP. CIP parameters, such as stride size and kernel size, matched those used in baseline comparisons. The input length was set to 96, and the future sequence was {96, 192, 336, 720}. Table 2 presents the results of the study. The reduced values of the MSE and MAE with added components indicate their respective substantial contributionsIn particular, the addition of ICM leads to further improvements, demonstrating the importance of modeling inter-variable interactions. 

\def\scaleSize {0.9}
\begin{center}
\begin{table}[th]
\begin{tabular}{cc|cc|cc|cc}

\toprule[1pt]
\multicolumn{2}{c|}{\scalebox{0.95}{Model}}& 
\multicolumn{2}{c|}{\scalebox{0.95}{W/O IConv}}& 
\multicolumn{2}{c|}{\scalebox{0.95}{W/O ICM}}& 
\multicolumn{2}{c}{\scalebox{0.95}{IConv}}
\\
\bottomrule[0.25pt]

\toprule[0.25pt]
\multicolumn{2}{c|}{\scalebox{0.9}{Metric}}                     
    &\scalebox{\scaleSize}{MSE}&\scalebox{\scaleSize}{MAE}
    &\scalebox{\scaleSize}{MSE}&\scalebox{\scaleSize}{MAE}
    &\scalebox{\scaleSize}{MSE}&\scalebox{\scaleSize}{MAE}
\\
\bottomrule[0.25pt]

\toprule[0.25pt]
\multicolumn{1}{c|}
{\multirow{4}{*}{\rotatebox[origin=c]{90}{Solar}}} 
    &\scalebox{\scaleSize}{96}  
        &\scalebox{\scaleSize}{0.236}&\scalebox{\scaleSize}{0.246}
        &\scalebox{\scaleSize}{0.221}&\scalebox{\scaleSize}{0.228}
        &\scalebox{\scaleSize}{\textbf{0.194}}&\scalebox{\scaleSize}{\textbf{0.209}}
                
    \\

    \multicolumn{1}{c|}{}
    &\scalebox{\scaleSize}{192}
        &\scalebox{\scaleSize}{0.286}&\scalebox{\scaleSize}{0.273}
        &\scalebox{\scaleSize}{0.259}&\scalebox{\scaleSize}{0.249}
        &\scalebox{\scaleSize}{\textbf{0.237}}&\scalebox{\scaleSize}{\textbf{0.232}}

    \\
    \multicolumn{1}{c|}{}                     
    &\scalebox{\scaleSize}{336}
        &\scalebox{\scaleSize}{0.317}&\scalebox{\scaleSize}{0.290}
        &\scalebox{\scaleSize}{0.287}&\scalebox{\scaleSize}{0.264}
        &\scalebox{\scaleSize}{\textbf{0.259}}&\scalebox{\scaleSize}{\textbf{0.245}}

    \\
    \multicolumn{1}{c|}{}                     
    &\scalebox{\scaleSize}{720}
        &\scalebox{\scaleSize}{0.318}&\scalebox{\scaleSize}{0.289}
        &\scalebox{\scaleSize}{0.290}&\scalebox{\scaleSize}{0.265}
        &\scalebox{\scaleSize}{\textbf{0.262}}&\scalebox{\scaleSize}{\textbf{0.248}}
\\
\bottomrule[0.25pt]

\toprule[0.25pt]
\multicolumn{1}{c|}
{\multirow{4}{*}{\rotatebox[origin=c]{90}{ECL}}} 
    &\scalebox{\scaleSize}{96}  
        &\scalebox{\scaleSize}{0.200}&\scalebox{\scaleSize}{0.276}
        &\scalebox{\scaleSize}{0.166}&\scalebox{\scaleSize}{0.250}
        &\scalebox{\scaleSize}{\textbf{0.140}}&\scalebox{\scaleSize}{\textbf{0.230}}
                
    \\

    \multicolumn{1}{c|}{}
    &\scalebox{\scaleSize}{192}
        &\scalebox{\scaleSize}{0.205}&\scalebox{\scaleSize}{0.284}
        &\scalebox{\scaleSize}{0.180}&\scalebox{\scaleSize}{0.261}
        &\scalebox{\scaleSize}{\textbf{0.156}}&\scalebox{\scaleSize}{\textbf{0.245}}

    \\
    \multicolumn{1}{c|}{}                     
    &\scalebox{\scaleSize}{336}
        &\scalebox{\scaleSize}{0.219}&\scalebox{\scaleSize}{0.298}
        &\scalebox{\scaleSize}{0.193}&\scalebox{\scaleSize}{0.275}
        &\scalebox{\scaleSize}{\textbf{0.167}}&\scalebox{\scaleSize}{\textbf{0.258}}

    \\
    \multicolumn{1}{c|}{}                     
    &\scalebox{\scaleSize}{720}
        &\scalebox{\scaleSize}{0.259}&\scalebox{\scaleSize}{0.329}
        &\scalebox{\scaleSize}{0.233}&\scalebox{\scaleSize}{0.308}
        &\scalebox{\scaleSize}{\textbf{0.194}}&\scalebox{\scaleSize}{\textbf{0.281}}

\\
\bottomrule[0.25pt]

\toprule[0.25pt]
\multicolumn{1}{c|}
{\multirow{4}{*}{\rotatebox[origin=c]{90}{Traffic}}} 
    &\scalebox{\scaleSize}{96}  
        &\scalebox{\scaleSize}{0.529}&\scalebox{\scaleSize}{0.336}
        &\scalebox{\scaleSize}{0.509}&\scalebox{\scaleSize}{0.305}
        &\scalebox{\scaleSize}{\textbf{0.459}}&\scalebox{\scaleSize}{\textbf{0.263}}
                
    \\

    \multicolumn{1}{c|}{}
    &\scalebox{\scaleSize}{192}
        &\scalebox{\scaleSize}{0.526}&\scalebox{\scaleSize}{0.336}
        &\scalebox{\scaleSize}{0.506}&\scalebox{\scaleSize}{0.302}
        &\scalebox{\scaleSize}{\textbf{0.466}}&\scalebox{\scaleSize}{\textbf{0.271}}

    \\
    \multicolumn{1}{c|}{}                     
    &\scalebox{\scaleSize}{336}
        &\scalebox{\scaleSize}{0.540}&\scalebox{\scaleSize}{0.340}
        &\scalebox{\scaleSize}{0.516}&\scalebox{\scaleSize}{0.305}
        &\scalebox{\scaleSize}{\textbf{0.479}}&\scalebox{\scaleSize}{\textbf{0.274}}

    \\
    \multicolumn{1}{c|}{}                     
    &\scalebox{\scaleSize}{720}
        &\scalebox{\scaleSize}{0.572}&\scalebox{\scaleSize}{0.575}
        &\scalebox{\scaleSize}{0.550}&\scalebox{\scaleSize}{0.322}
        &\scalebox{\scaleSize}{\textbf{0.517}}&\scalebox{\scaleSize}{\textbf{0.293}}

\\
\bottomrule[1pt]

\end{tabular}
\caption{Results of ablation study of each component on Solar Energy, ECL, and Traffic datasets}
\end{table}
\end{center}

\begin{figure}[th]
\centering
\includegraphics[width=0.95\columnwidth, trim = 0cm 1cm 0cm 0cm, clip]{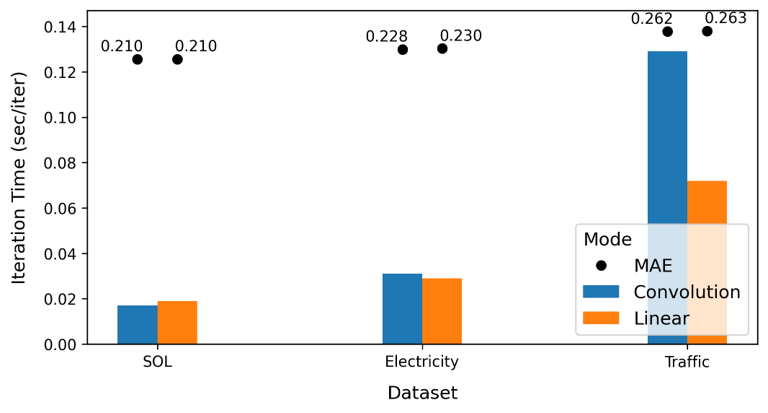} 
\caption{Model efficiency comparison between convolutional and linear layers}
\label{fig5}
\end{figure}

\subsection{Comparison of Convolution and Linear layer }
We validated the replacement of the $1\times 1$ convolution layer with a linear layer through comparative experiments, as illustrated in Figure 5. The weights of each layer were initialized by $N(0,0.01)$. Despite implementation differences, both models produced similar outputs. While the convolutional layer demonstrated reduced processing time for smaller datasets, the linear layer’s processing time starts outperform the convolutional layer as the number of variables increased. For the traffic dataset, the linear layer performed 45\% faster than the convolutional layer.

\subsection{Efficiency Analysis}

A comparative analysis of the model's computational efficiency, considering training speed and memory consumption, is presented in Figure 6. Training speed was quantified by calculating the mean inference time. The necessary memory is calculated based on the model's parameter size. The figure indicates that IConv achieved substantial memory reductions compared to other convolution-based MTSF models across datasets with both large and small variable counts. In large variable datasets, a considerable reduction in computation time is observed, outperforming transformer-based models.

\begin{figure}[th]
\centering
\includegraphics[width=0.95\columnwidth]{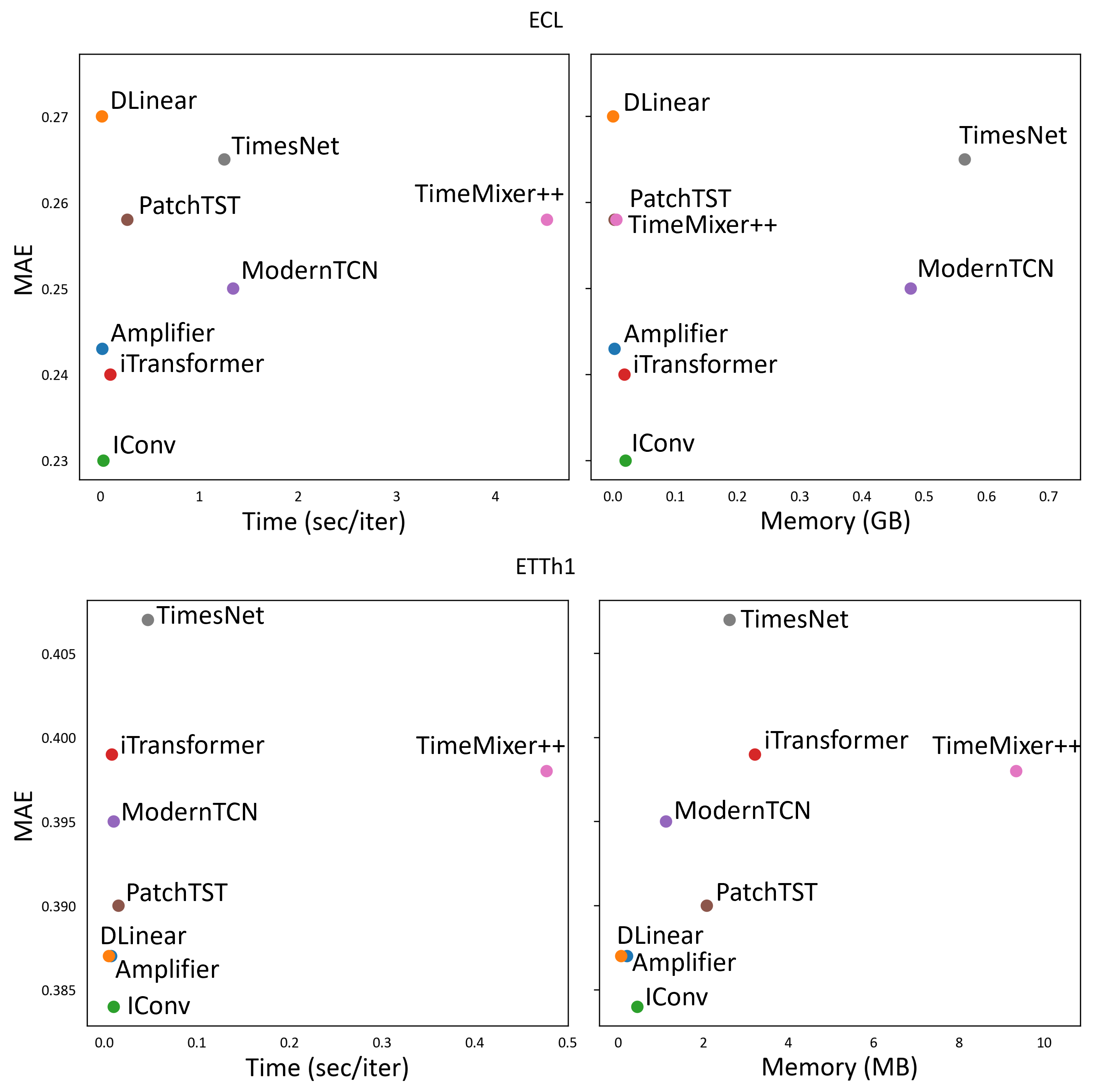} 
\caption{Comparison of computation time and prediction performance between convolution and linear layers}
\label{fig6}
\end{figure}

\section{Conclusion}
This paper introduced a novel hybrid MLP-CNN model for time series forecasting, addressing limitations of existing MLP-based approaches. Our model effectively captures long-term dependencies using MLPs and local variations via a specialized convolutional network, IConv. By decoupling temporal and inter-channel processing, we significantly improved efficiency and predictive accuracy, achieving state-of-the-art performance on multiple datasets while reducing computational costs compared to existing convolution-based models. Moreover, the visualization of prediction and receptive field has allowed us to validate the hypothesis concerning the efficacy of IConv in discerning local temporal patterns.

\section{Appendix}

\section{1 Experimentation Details}

\subsection{1.1 Datasets}
We evaluated IConv's performance on six real-world datasets for a comparative analysis. A detailed description of each dataset is provided below: (1) ECL (Electricity Consuming Load): This dataset \cite{Autoformer} contains hourly electricity consumption data for 321 clients. (2) ETT (Electricity Transformer Temperature): From July 2016 to July 2018, this dataset \cite{Autoformer} records seven factors related to electricity transformers. It is divided into four subsets: ETTh1 and ETTh2 (hourly data), and ETTm1 and ETTm2 (15-minute data). (3) Solar-Energy: This dataset \cite{LTSNet} records solar power production from 137 photovoltaic (PV) plants in 2006, with measurements taken every 10 minutes. (4) Traffic: The traffic dataset \cite{Autoformer} contains hourly road occupancy rates from 862 sensors on San Francisco Bay area freeways, collected from January 2015 to December 2016. (5) Weather: The weather dataset \cite{Autoformer} includes 21 meteorological factors collected every 10 minutes throughout 2020 from the Weather Station of the Max Planck Biogeochemistry Institute. A summary of these datasets is presented in Table 3.

\def\scaleSize {0.9}
\begin{center}
\begin{table}[th]
\renewcommand{\arraystretch}{1.2}
\begin{tabular}{c|ccc}

\hline
\scalebox{0.95}{Model}& 
\scalebox{0.95}{Dim}& 
\scalebox{0.95}{\makecell{Dataset size \\  (Train, Val, Test)}}& 
\scalebox{0.95}{Freq}
\\
\hline
ECL & 321 & (18317, 2633, 5261) & Hourly \\
ETTm & 7 & (34465, 11521, 11521) & 15 mins \\
ETTh & 7 & (8545, 2881, 2881) & Hourly \\
Solar Energy & 127 & (36601, 5161, 10417) & 10 mins \\
Traffic & 862 & (12185, 1757, 3509) & Hourly \\
Weather & 21 & (36792, 5271, 10540) & 10 mins \\
\hline

\end{tabular}
\caption{Summary of dataset}
\end{table}
\end{center}

\subsection{1.2 Model Architecture}
While the main paper presents the core equations and figures for a single IConv layer, this appendix provides a more detailed explanation of the overall model architecture, which consists of multiple IConv layers. The model employs a sequential structure where three IConv operations are performed subsequently. The detailed order of each operation is explained through Algorithm 1. This design allows the model to extract progressively finer details from the overall trend $\tilde{Y}$. In our experiments, we maintained consistent stride and multiplier values across all IConv layers, such that $M_1=M_2=M_3$ and $S_1=S_2=S_3$. The patch sizes, however, were configured to decrease sequentially, satisfying the condition $P_1>P_2>P_3$. The detailed setting related with kernel size and stride size is explained in next subsection. 

\begin{algorithm}[ht]
	\caption{MLP-IConv Architecture}
	\label{MLP-IConv Architecture}
	\begin{algorithmic}[1]
	    \renewcommand{\algorithmicrequire}{\textbf{Input:}}
		\REQUIRE $X_{in}\in R^{C\times T}$
        \renewcommand{\algorithmicensure}{\textbf{Parameters:}}
		\ENSURE $\newline P=[P_1,P_2,P_3] \newline S=[S_1,S_2,S_3] \newline M=[M_1,M_2,M_3]$
	    \renewcommand{\algorithmicensure}{\textbf{Output:}}
		\ENSURE $\hat{Y}\in R^{C\times L}$
		\STATE $X_{enc} = Enc(X_{in})$
		\STATE $\tilde{Y} = Reg(X_{enc})$
		\FOR {$i=0$ to $len(P)$}
			\STATE $H=CIPC(\tilde{Y},P[i],S[i],M[i])$
            \STATE $\hat{H} = ICM(H)$
            \STATE $V=CIPE(\hat{H},P[i],S[i],M[i])$
            \STATE $\tilde{Y} = \tilde{Y} + Var(\tilde{Y}) \times V$
        \ENDFOR
        \STATE $\hat{Y} = \tilde{Y}$
        \RETURN $\hat{Y}$
	\end{algorithmic}
\end{algorithm}

\subsection{1.3 Hyperparameter Sensitivity}
To explore the impact of different configurations, we tested various combinations of kernel and stride sizes. For our experiments, we restricted the stride sizes to either 3 or 4. To help the model better capture the underlying patterns of the datasets, we matched the stride size to the time interval of the dataset. For instance, the solar energy dataset, which is sampled every 10 minutes, uses a stride size of 3 to analyze hourly temporal patterns (10 minutes$\times$3=30 minutes). Following this rule, we explored the best kernel sizes from {36,24,12}, {18,12,6}, and {12,6,3} for the solar energy dataset. Similarly, for the ECL and Traffic da-tasets, we experimented with kernel sizes from {36,24,12}, {24,16,8}, and {12,8,4} with a stride size of 4. The results of these experiments are summarized in Tables 3 and 4, with a visual comparison provided in Figure 7. Our findings suggest that a larger multiplier generally improves the model's predictive accuracy by enabling the extraction of a more diverse representation. In contrast, a larger kernel size does not consistently lead to better performance. The optimal kernel size appears to vary significantly depending on the dataset's specific characteristics and the desired prediction length. This indicates a potential performance boost when the data exhibits evident periodicity, as the model can better exploit these patterns.

\def\scaleSize {1.0}
\begin{table*}[!htb]
\begin{center}
\resizebox{0.95\textwidth}{!}{%
\begin{tabular}{c|c|cccc|cccc|cccc}

\toprule[1pt]
\multicolumn{2}{c|}{\scalebox{0.95}{Kernel size set}}& 
\multicolumn{4}{c|}{\scalebox{0.95}{\{12,6,3\}}}& 
\multicolumn{4}{c|}{\scalebox{0.95}{\{18,12,6\}}}& 
\multicolumn{4}{c}{\scalebox{0.95}{\{36,24,12\}}}

\\
\bottomrule[0.25pt]

\toprule[0.25pt]
\multicolumn{2}{c|}{\scalebox{0.9}{Multiplier}}                     
    &\scalebox{\scaleSize}{3}&\scalebox{\scaleSize}{4}&\scalebox{\scaleSize}{6}&\scalebox{\scaleSize}{8}
    &\scalebox{\scaleSize}{3}&\scalebox{\scaleSize}{4}&\scalebox{\scaleSize}{6}&\scalebox{\scaleSize}{8}
    &\scalebox{\scaleSize}{3}&\scalebox{\scaleSize}{4}&\scalebox{\scaleSize}{6}&\scalebox{\scaleSize}{8}
\\
\bottomrule[0.25pt]

\toprule[0.25pt]
\multicolumn{1}{c|}
{\multirow{4}{*}{\rotatebox[origin=c]{90}{Solar}}} 
    &\scalebox{\scaleSize}{96} 
        &\scalebox{\scaleSize}{0.212}&\scalebox{\scaleSize}{0.212}&\scalebox{\scaleSize}{0.213}&\scalebox{\scaleSize}{0.212}
        &\scalebox{\scaleSize}{0.213}&\scalebox{\scaleSize}{0.210}&\scalebox{\scaleSize}{\underline{0.208}}&\scalebox{\scaleSize}{0.209}
        &\scalebox{\scaleSize}{0.210}&\scalebox{\scaleSize}{0.209}&\scalebox{\scaleSize}{0.209}&\scalebox{\scaleSize}{\textbf{0.206}}
    \\
    &\scalebox{\scaleSize}{192}
        &\scalebox{\scaleSize}{0.233}&\scalebox{\scaleSize}{0.235}&\scalebox{\scaleSize}{0.232}&\scalebox{\scaleSize}{\textbf{0.229}}
        &\scalebox{\scaleSize}{0.232}&\scalebox{\scaleSize}{\underline{0.230}}&\scalebox{\scaleSize}{0.235}&\scalebox{\scaleSize}{\textbf{0.229}}
        &\scalebox{\scaleSize}{0.240}&\scalebox{\scaleSize}{0.235}&\scalebox{\scaleSize}{0.233}&\scalebox{\scaleSize}{0.231}
    \\
    &\scalebox{\scaleSize}{336}
        &\scalebox{\scaleSize}{0.248}&\scalebox{\scaleSize}{0.252}&\scalebox{\scaleSize}{0.248}&\scalebox{\scaleSize}{\textbf{0.245}}
        &\scalebox{\scaleSize}{\textbf{0.245}}&\scalebox{\scaleSize}{\textbf{0.245}}&\scalebox{\scaleSize}{0.248}&\scalebox{\scaleSize}{\underline{0.246}}
        &\scalebox{\scaleSize}{0.257}&\scalebox{\scaleSize}{0.248}&\scalebox{\scaleSize}{0.250}&\scalebox{\scaleSize}{0.247}
    \\
    &\scalebox{\scaleSize}{720}
        &\scalebox{\scaleSize}{0.251}&\scalebox{\scaleSize}{0.251}&\scalebox{\scaleSize}{0.250}&\scalebox{\scaleSize}{0.250}
        &\scalebox{\scaleSize}{\textbf{0.247}}&\scalebox{\scaleSize}{0.250}&\scalebox{\scaleSize}{0.251}&\scalebox{\scaleSize}{\underline{0.249}}
        &\scalebox{\scaleSize}{0.255}&\scalebox{\scaleSize}{\textbf{0.247}}&\scalebox{\scaleSize}{0.250}&\scalebox{\scaleSize}{0.250}
		
\\

\bottomrule[1pt]

\end{tabular}
}
\end{center}
\caption{Hyperparameter sensitivity results on the ECL and traffic datasets with stride size 4 measured by MAE}
\end{table*}

\def\scaleSize {1.0}
\begin{table*}[!htb]
\begin{center}
\resizebox{0.95\textwidth}{!}{%
\begin{tabular}{c|c|cccc|cccc|cccc}

\toprule[1pt]
\multicolumn{2}{c|}{\scalebox{0.95}{Kernel size set}}& 
\multicolumn{4}{c|}{\scalebox{0.95}{\{12,8,4\}}}& 
\multicolumn{4}{c|}{\scalebox{0.95}{\{24,16,8\}}}& 
\multicolumn{4}{c}{\scalebox{0.95}{\{36,24,12\}}}

\\
\bottomrule[0.25pt]

\toprule[0.25pt]
\multicolumn{2}{c|}{\scalebox{0.9}{Multiplier}}                     
    &\scalebox{\scaleSize}{3}&\scalebox{\scaleSize}{4}&\scalebox{\scaleSize}{6}&\scalebox{\scaleSize}{8}
    &\scalebox{\scaleSize}{3}&\scalebox{\scaleSize}{4}&\scalebox{\scaleSize}{6}&\scalebox{\scaleSize}{8}
    &\scalebox{\scaleSize}{3}&\scalebox{\scaleSize}{4}&\scalebox{\scaleSize}{6}&\scalebox{\scaleSize}{8}
\\
\bottomrule[0.25pt]

\toprule[0.25pt]
\multicolumn{1}{c|}
{\multirow{4}{*}{\rotatebox[origin=c]{90}{ECL}}} 
    &\scalebox{\scaleSize}{96} 
        &\scalebox{\scaleSize}{0.234}&\scalebox{\scaleSize}{0.232}&\scalebox{\scaleSize}{\underline{0.230}}&\scalebox{\scaleSize}{\textbf{0.229}}
        &\scalebox{\scaleSize}{0.233}&\scalebox{\scaleSize}{0.232}&\scalebox{\scaleSize}{0.231}&\scalebox{\scaleSize}{\underline{0.230}}
        &\scalebox{\scaleSize}{0.234}&\scalebox{\scaleSize}{0.233}&\scalebox{\scaleSize}{0.232}&\scalebox{\scaleSize}{\underline{0.230}}
    \\
    &\scalebox{\scaleSize}{192}
        &\scalebox{\scaleSize}{0.246}&\scalebox{\scaleSize}{0.246}&\scalebox{\scaleSize}{\underline{0.244}}&\scalebox{\scaleSize}{\textbf{0.243}}
        &\scalebox{\scaleSize}{0.247}&\scalebox{\scaleSize}{0.246}&\scalebox{\scaleSize}{\underline{0.244}}&\scalebox{\scaleSize}{0.245}
        &\scalebox{\scaleSize}{0.248}&\scalebox{\scaleSize}{0.248}&\scalebox{\scaleSize}{0.246}&\scalebox{\scaleSize}{0.246}
    \\
    &\scalebox{\scaleSize}{336}
        &\scalebox{\scaleSize}{0.261}&\scalebox{\scaleSize}{0.260}&\scalebox{\scaleSize}{\textbf{0.257}}&\scalebox{\scaleSize}{\textbf{0.257}}
        &\scalebox{\scaleSize}{0.260}&\scalebox{\scaleSize}{0.259}&\scalebox{\scaleSize}{\underline{0.258}}&\scalebox{\scaleSize}{\underline{0.258}}
        &\scalebox{\scaleSize}{0.261}&\scalebox{\scaleSize}{0.260}&\scalebox{\scaleSize}{0.259}&\scalebox{\scaleSize}{\textbf{0.257}}
    \\
    &\scalebox{\scaleSize}{720}
        &\scalebox{\scaleSize}{0.284}&\scalebox{\scaleSize}{0.283}&\scalebox{\scaleSize}{\underline{0.282}}&\scalebox{\scaleSize}{\textbf{0.281}}
        &\scalebox{\scaleSize}{0.286}&\scalebox{\scaleSize}{0.283}&\scalebox{\scaleSize}{\textbf{0.281}}&\scalebox{\scaleSize}{\textbf{0.281}}
        &\scalebox{\scaleSize}{0.284}&\scalebox{\scaleSize}{0.284}&\scalebox{\scaleSize}{0.283}&\scalebox{\scaleSize}{0.283}
\\
\bottomrule[0.25pt]

\toprule[0.25pt]
\multicolumn{1}{c|}
{\multirow{4}{*}{\rotatebox[origin=c]{90}{Traffic}}} 
    &\scalebox{\scaleSize}{96} 
        &\scalebox{\scaleSize}{0.276}&\scalebox{\scaleSize}{0.272}&\scalebox{\scaleSize}{0.270}&\scalebox{\scaleSize}{0.267}
        &\scalebox{\scaleSize}{0.275}&\scalebox{\scaleSize}{0.270}&\scalebox{\scaleSize}{0.265}&\scalebox{\scaleSize}{\underline{0.263}}
        &\scalebox{\scaleSize}{0.274}&\scalebox{\scaleSize}{0.270}&\scalebox{\scaleSize}{0.266}&\scalebox{\scaleSize}{\textbf{0.262}}
    \\
    &\scalebox{\scaleSize}{192}
        &\scalebox{\scaleSize}{0.280}&\scalebox{\scaleSize}{0.276}&\scalebox{\scaleSize}{0.273}&\scalebox{\scaleSize}{\underline{0.271}}
        &\scalebox{\scaleSize}{0.278}&\scalebox{\scaleSize}{0.275}&\scalebox{\scaleSize}{\underline{0.271}}&\scalebox{\scaleSize}{\underline{0.271}}
        &\scalebox{\scaleSize}{0.279}&\scalebox{\scaleSize}{0.275}&\scalebox{\scaleSize}{0.272}&\scalebox{\scaleSize}{\textbf{0.270}}
    \\
    &\scalebox{\scaleSize}{336}
        &\scalebox{\scaleSize}{0.283}&\scalebox{\scaleSize}{0.279}&\scalebox{\scaleSize}{0.278}&\scalebox{\scaleSize}{0.276}
        &\scalebox{\scaleSize}{0.281}&\scalebox{\scaleSize}{0.279}&\scalebox{\scaleSize}{0.276}&\scalebox{\scaleSize}{\textbf{0.274}}
        &\scalebox{\scaleSize}{0.282}&\scalebox{\scaleSize}{0.279}&\scalebox{\scaleSize}{0.276}&\scalebox{\scaleSize}{\underline{0.275}}
    \\
    &\scalebox{\scaleSize}{720}
        &\scalebox{\scaleSize}{0.300}&\scalebox{\scaleSize}{0.298}&\scalebox{\scaleSize}{0.295}&\scalebox{\scaleSize}{0.297}
        &\scalebox{\scaleSize}{0.300}&\scalebox{\scaleSize}{0.296}&\scalebox{\scaleSize}{0.295}&\scalebox{\scaleSize}{\textbf{0.293}}
        &\scalebox{\scaleSize}{0.299}&\scalebox{\scaleSize}{0.297}&\scalebox{\scaleSize}{\underline{0.294}}&\scalebox{\scaleSize}{0.295}		

\\

\bottomrule[1pt]

\end{tabular}
}

\end{center}
\caption{Hyperparameter sensitivity results on the solar energy dataset with stride size 3 measured by MAE}
\end{table*}

\begin{figure*}[!htb]
\centering
\includegraphics[width=0.95\textwidth]{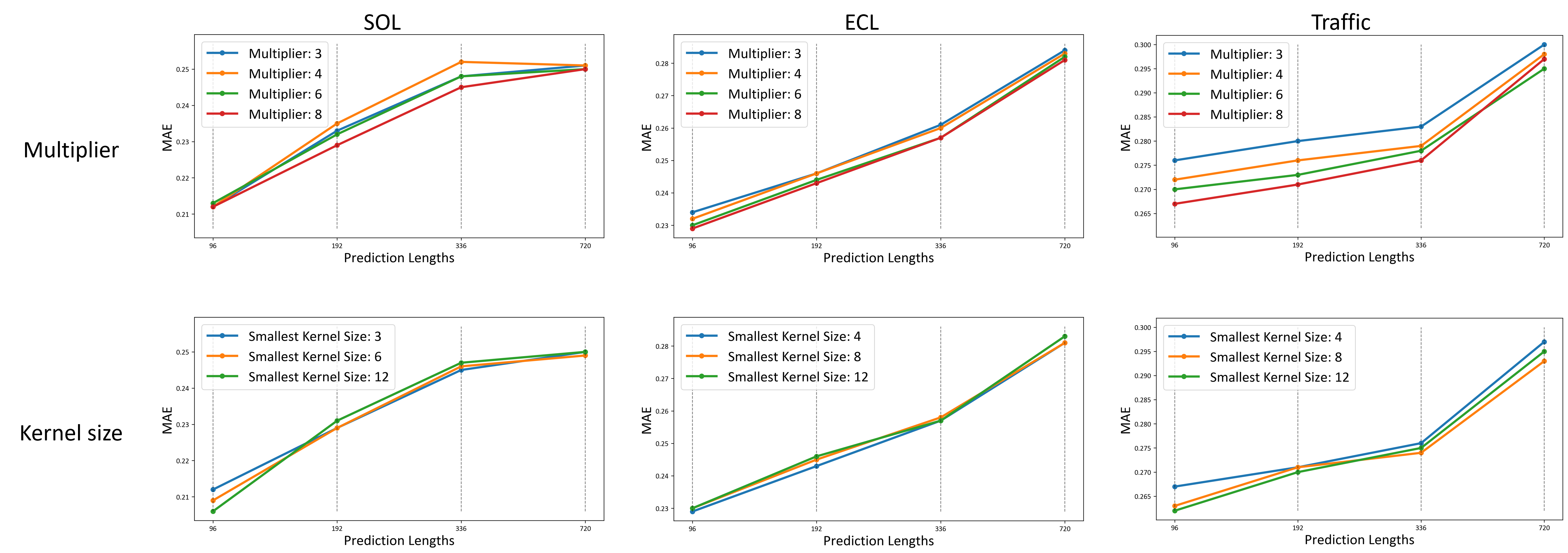} 
\caption{Visualization of hyperparameter sensitivity for each multiplier and kernel size}
\label{fig7}
\end{figure*}

\section{2 Features of IConv}
\subsection{2.1 Receptive Field}
To analyze the receptive field of our IConv model, we adopted the methodology of Luo and Wang (2024). We randomly sampled 50 data instances from the validation set and computed their outputs, denoted as $\hat{Y} \in R^{50\times C\times L}$. To represent the relationship between the input and output timesteps, we averaged the batch-wise and channel-wise dimensions of the prediction sequence's midpoint. This is expressed by the following equations:

\begin{equation}
F=\frac{1}{C}\frac{1}{50} \sum_{c=1}^{C} \sum_{i=1}^{50} \hat{Y}[i,L/2,c]
\end{equation}
\begin{equation}
X_G=\frac{1}{C}\frac{1}{50} \sum_{c=1}^{C} \sum_{i=1}^{50} X_in[i,:,c]
\end{equation}
Next, we calculated the gradient of F with respect to each averaged input timestep, represented as:
\begin{equation}
G=\frac{\partial F}{\partial X_G^{(t)}}
\end{equation}

The computed gradient values were then normalized to the range [0,1] using min-max scaling. These results are visualized as a heatmap in Figure 4 of the main paper. In this appendix, heatmaps for datasets other than ECL are presented in Figure 8. As shown in Figure 8(c), the receptive field for the Traffic dataset mirrors that of the ECL dataset, indicating that the model utilizes contextual information surrounding the individual timestep. For the weather dataset, represented in Figure 8(a), the receptive field is visibly extended compared to traditional MLP models. The visualization of the receptive field clearly demonstrates IConv's efficacy in modeling local temporal patterns from the input sequence. However, this approach does not fully explain the relationship between the input sequence and the predicted output sequence in MLP models. For a more comprehensive understanding of which timesteps are referenced for prediction, a visualization of the model's weight matrix is a more suitable approach.

\subsection{2.2 Weight Matrix}

While receptive field visualization provides insight into how IConv processes the input sequence, a weight matrix visualization offers a deeper understanding of its contribution to the predicted output. To validate IConv's efficacy, we have visualized the weight matrices of from three models: regression layer of IConv ($W_reg$), a standard MLP, and the trend component of DLinear. The results are presented in Figure 9. The values along the main diagonal of these matrices reflect the models' ability to capture evident periodicity and overall trends within the time-series data. In the weight matrix of weather and solar energy datasets, depicted in Figures 9(c1) and 9(d1), the diagonal line is notably more pronounced than in both the MLP and DLinear matrices. Additionally, the IConv weight matrix demonstrates a more restrained and less probabilistic distribution of values in the non-diagonal areas. This is even more apparent in the ECL and Traffic datasets. The IConv weight matrix, as shown in Figures 9(b1) and 9(e1), demonstrates a strong diagonal pattern, prioritizing the capture of global periodicity. In contrast, the non-diagonal areas of the MLP and DLinear weight matrices exhibit higher activation, suggesting a greater reliance on less-structured information. These findings highlight how IConv's architectural design encourages the regression layer to focus on capturing global patterns, while its convolutional components effectively manage local temporal variations.

\begin{figure*}[!htb]
\centering
\includegraphics[width=0.95\textwidth]{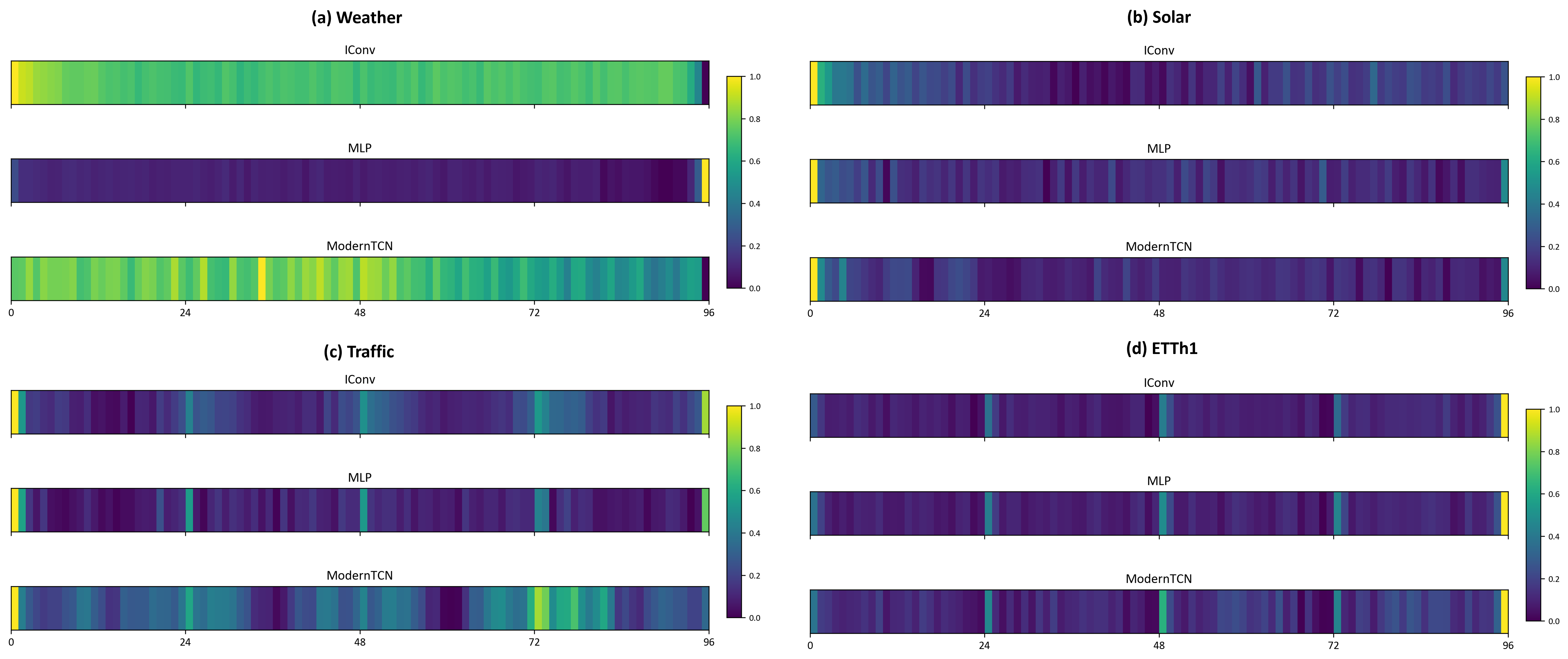} 
\caption{Visualization of receptive field for (a) weather, (b) solar energy, (c) traffic, and (d) ETTh1 dataset}
\label{fig8}
\end{figure*}

\begin{figure*}[!htb]
\centering
\includegraphics[width=0.95\textwidth]{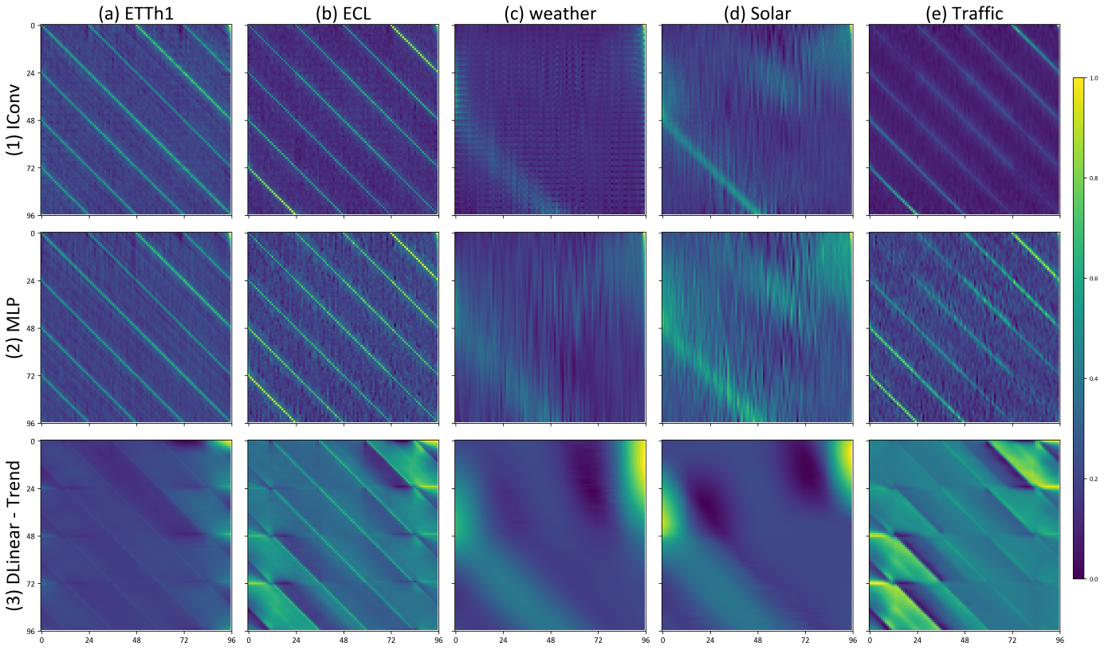} 
\caption{Visualization of weight matrix for MLP and IConv, where the index within the grid is specified in the bracket}
\label{fig9}
\end{figure*}

\pagebreak

\section{Acknowledgements}
This research was supported by Korea Institute of Marine Science \& Technology Promotion (KIMST) funded by the Ministry of Oceans and Fisheries (RS-2022-KS221657).

\bibliography{IConv}

\end{document}